\documentclass[lettersize,journal]{IEEEtran}
\usepackage{tikz}
\usepackage{multirow}
\usepackage{graphicx}
\usepackage{booktabs}
\usepackage{siunitx}
\usepackage{pifont}
\usepackage{enumitem}
\usepackage{amsmath,amsfonts}
\usepackage{algorithmic}
\usepackage{algorithm}
\usepackage{array}
\usepackage[caption=false,font=normalsize,labelfont=sf,textfont=sf]{subfig}
\usepackage{textcomp}
\usepackage{stfloats}
\usepackage{url}
\usepackage{verbatim}
\usepackage{graphicx}
\usepackage{cite}
\begin{document}
\title{SeFENet: Robust Deep Homography Estimation via  \\ Semantic-Driven Feature Enhancement}

\author{Zeru~Shi,~Zengxi~Zhang,~Kemeng~Cui,~Ruizhe~An,~Jinyuan~Liu~\IEEEmembership{Member,~IEEE}~and~Zhiying~Jiang$^{*}$,
	
	\thanks{Zeru Shi, Zengxi Zhang, Kemeng Cui and Ruizhe An are with the School of Software Technology, Dalian University of Technology, Dalian 116024, China (e-mail: shizeru77@gmail.com, cyouzoukyuu@gmail.com, giovanna1230@163.com, anruizhe1@outlook.com;).}
	\thanks{Jinyuan Liu is with the School of Mechanical Engineering, Dalian University of Technology, Dalian 116024, China (e-mail: atlantis918@hotmail.com).}
        \thanks{Zhiying Jiang is with the College of Information Science and Technology, Dalian Maritime University, Dalian 116026, China (e-mail: zyjiang0630@gmail.com).}
        \thanks{Corresponding author: Zhiying Jiang.}
}
\markboth{Journal of \LaTeX\ Class Files,~Vol.~14, No.~8, August~2021}%
{Shell \MakeLowercase{\textit{et al.}}: A Sample Article Using IEEEtran.cls for IEEE Journals}


\maketitle
\begin{abstract}
Images captured in harsh environments often exhibit blurred details, reduced contrast, and color distortion, which hinder feature detection and matching, thereby affecting the accuracy and robustness of homography estimation. While visual enhancement can improve contrast and clarity, it may introduce visual-tolerant artifacts that obscure the structural integrity of images. Considering the resilience of semantic information against environmental interference, we propose a semantic-driven feature enhancement network for robust homography estimation, dubbed SeFENet. Concretely, we first introduce an innovative hierarchical scale-aware module to expand the receptive field by aggregating multi-scale information, thereby effectively extracting image features under diverse harsh conditions. Subsequently, we propose a semantic-guided constraint module combined with a high-level perceptual framework to achieve degradation-tolerant with semantic feature. A meta-learning-based training strategy is introduced to mitigate the disparity between semantic and structural features. By internal-external alternating optimization, the proposed network achieves implicit semantic-wise feature enhancement, thereby improving the robustness of homography estimation in adverse environments by strengthening the local feature comprehension and context information extraction. Experimental results under both normal and harsh conditions demonstrate that SeFENet significantly outperforms SOTA methods, reducing point match error by at least 41\% on the large-scale datasets.
\end{abstract}


\section{Introduction}
\IEEEPARstart{H}{omography} is a global projective transformation that translates points captured from various viewpoints in one image to corresponding points in another image. It has extensive applications in computer vision, spanning from monocular camera systems to multi-camera systems. These applications include image and video stitching~\cite{brown2007automatic,gao2011constructing,guo2016joint}, multi-scale gigapixel photography~\cite{brady2012multiscale,shao2021localtrans}, multispectral image fusion~\cite{ying2021unaligned, zhou2019integrated}, planar object tracking~\cite{zhan2022homography, zhang2022hvc}, simultaneous localization and mapping (SLAM)~\cite{mur2015orb, engel2014lsd}, and UAV localization in GPS-denied environment~\cite{goforth2019gps, zhao2021deep}.
\begin{figure}[t]
    \centering
    \includegraphics[width=0.48\textwidth]{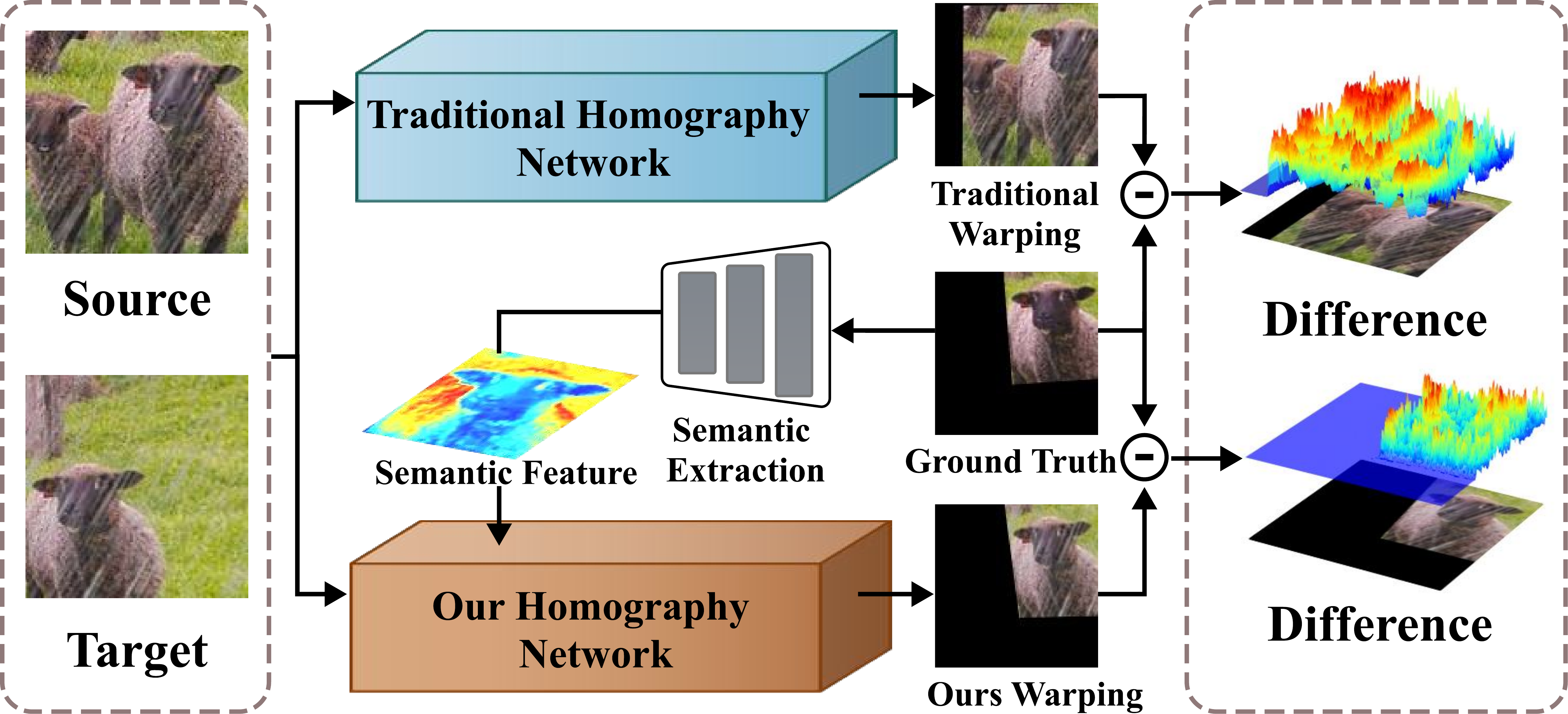}
    \caption{The concept diagram of this paper is shown above. The top half is other homography estimation methods and the bottom half is our homography estimation method. After integrating the semantics, it can be seen from the 3D topographic map that the distorted image of our homography estimation is closer to the source image than other methods in harsh environments.}
    \label{fig0:head}
\end{figure}

In the field of deep homography estimation, the pioneering approach~\cite{detone2016deep} utilized VGG-style networks to estimate the homography between concatenated image pairs. Building on this foundational framework, subsequent research~\cite{erlik2017homography,le2020deep, zhou2019integrated} introduced improvement by modifying network architectures or cascading multiple similar networks to improve accuracy. HomoGAN~\cite{zhan2022homography} generates a dominant plane mask with explicit coplanarity constraint to guide homography estimation to focus on the dominant plane, which significantly reduces the error. CAHomo~\cite{zhang2020content} learns a content-aware mask during the estimation process to reject outliers and employs a triple loss for unsupervised training. In addition, the work like MCNet~\cite{zhu2024mcnet} aims at effective iteration by integrating the iterative correlation search with a multi-scale strategy. Different from above works which focus on structural features, MaskHomo~\cite{Wang_Liu_Zhang_Xu_Wang_2024} introduces a homography estimation framework that leverages semantic information to effectively address issues related to multi-plane depth disparity. Although the methods above generally perform well under normal conditions, their accuracy declines significantly in harsh environments (e.g., rain, haze, and low light). These harsh conditions cause blurred details, information loss, and reduced contrast of captured images, thereby limiting the traditional structural feature-based methods on effective extraction and matching of features, especially under the large baseline condition~\cite{jiang2023semi}, which further imposes inevitable interference. While some works~\cite{zhang2020content, jiang2023supervised} enrich training with low-light images, they do not address the degradation of images in the network structure and training method, limiting their performance in challenging conditions.

In this paper, considering the limitation of existing methods in capturing information from images in harsh environments, we introduce a hierarchical scale-aware mechanism to increase the receptive fields. So as to focus on more contextual information and understand the relationship of objects in the image. Meanwhile, with the multi-scale structure feature, we combine them with multi-scale semantic features and propose a method named SeFENet, which enhances structural features by semantic features, which make homography network concerns more detailed information in the homography estimation in semantic-driven module, so that the network can better deal with the edge information and achieve robust homography estimation under either normal environment or harsh environments. Different from structure information, semantic features capture high-level information to emphasize the details of objects and ignore the interference in the background caused by harsh conditions. Thus, we leverage the module designed for high-level perception tasks to guide the proposed network in extracting effective semantic features from degraded images, thereby achieving robust estimation. Furthermore, since the large gap between semantic and structural features, we introduce a Semantic-Guide Meta Constrains module optimized by a meta-learning strategy to mitigate the feature gap, ensuring they jointly enhance the model's effectiveness. 
To summarize, our contributions are as follows:
 
\begin{itemize}
\item We propose SeFENet, a method that Implicitly enhances the semantic feature by the guidance of high-level perception tasks, resulting in achieving robust homography estimation in harsh and large-baseline environments.
\item A semantic-guide meta constraint is proposed to harmonize the semantic feature to the structure feature, preventing the negative interference caused by the gap between two different-level tasks.
\item We propose a hierarchical scale-aware feature extraction module that enhances the model's ability to capture multi-scale contextual information within an image scene. By expanding the receptive field during feature extraction, our approach effectively preserves essential details while filtering out irrelevant background noise, leading to more robust and accurate representations.
\item By experiment, SeFENet has demonstrated superior properties for homography estimation in harsh environment on both synthetic and real datasets.
\end{itemize}

\begin{figure*}[h]
    \centering
    \includegraphics[width=1\textwidth]{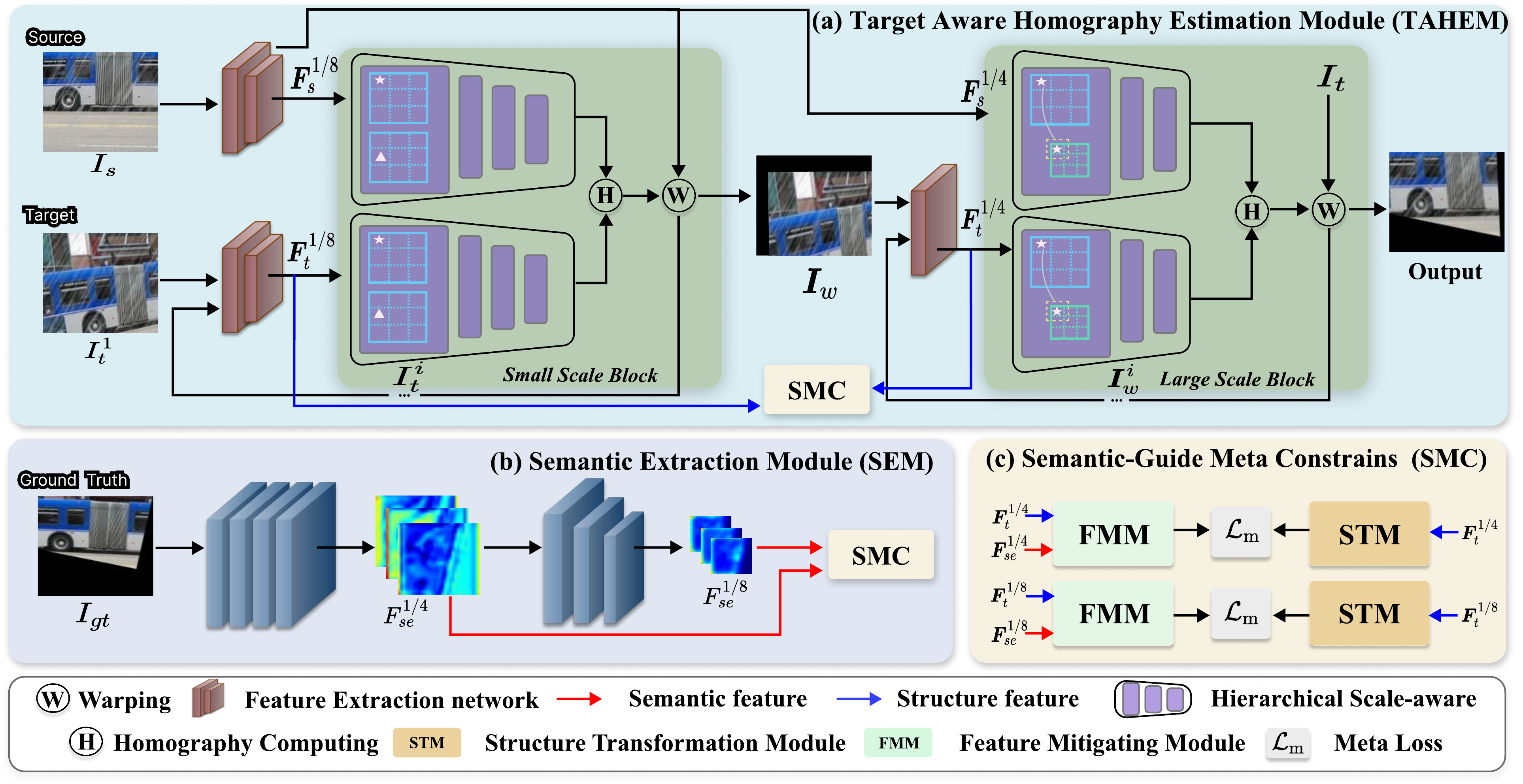}
    \caption{The overall pipeline of SeFENet. (a) is the detailed structure of target aware homography estimation module (TAHEM). (b) is the semantic extraction module (SEM). (c) is the structure of our semantic-guide meta constrains (SMC). Among the above modules, only (a) is used for inference, and the other modules are only used for the above processing during training.}
    \label{fig1:pipeine}
\end{figure*}

\section{Related Work}
\subsection{Homography Estimation}
Traditional homography estimation includes 3 key steps: feature extraction, feature matching and homography estimation. Common feature extraction methods include SIFT~\cite{lowe2004sift}, SURF~\cite{bay2006surf}, ORB~\cite{rublee2011orb}, GMS~\cite{bian2017gms}, BEBLID~\cite{suarez2020beblid}, SuperPoint~\cite{detone2018superpoint}. There are also deep learning-based methods for feature matching, such as LoFTR~\cite{sun2021loftr} and SuperGlue~\cite{sarlin2020superglue}. Popular homography estimation methods include RANSAC~\cite{derpanis2010overview} and MAGSAC~\cite{barath2019magsac}. However, these methods all rely on feature point detection, so the harsh acquisition environment will have a huge impact on the effect of homography estimation.

Now many methods abandon the traditional feature point matching and use higher level networks to extract features.~\cite{zhang2020content} introduces an approach which learns an outlier mask to identify and select only the most reliable regions for homography estimation without directly comparing image content and calculates the loss based on learned deep features. Unsupervised methods like~\cite{lu2022video} uses transformer to extract the features and calculate the correlation.~\cite{ye2021motion} proposes Feature Identity Loss (FIL) to enforce the learned image feature warp-equivariant. When learning homography without constraints,~\cite{hong2022unsupervised} addresses the issue of plane-induced parallax. LocalTrans proposed by~\cite{shao2021localtrans} addresses the cross-resolution issue in homography estimation. RealSH~\cite{jiang2023supervised} enhances the performance using an iterative strategy and the quality of the dataset.In addition to using structural features~\cite{liu2024semanticawarerepresentationlearninghomography, Wang_Liu_Zhang_Xu_Wang_2024} introduce semantic feature into homography estimation. However, when these methods face the image degradation in harsh environment, the matching results will appear double shadow, blur and so on.

\subsection{Meta-Learning}
Today, meta-learning has been applied in many fields, \cite{Wang_2019_ICCV, Elsken_2020_CVPR}. ~\cite{Volpi_2021_CVPR} proposes a meta-learning approach to create models that are less easy to drift, which is called catastrophic forgetting, when trained on different datasets constantly. A revised meta-learning framework proposed by~\cite{Min_2023_WACV} allows pre-trained networks to learn to adapt to new domains and enhance cross-domain performance.~\cite{zhang2024metadiff} introduces a new conditional diffusion-based meta-learning called MetaDiff. The optimization procedure of base learner weights from Gaussian initialization to target weights is modeled by MetaDiff in a denoising way. ~\cite{konwer2023enhancing} employs a meta-learning strategy during training to learn enhanced modality-agnostic representations even when only limited omnimodal samples are available.~\cite{wang2023improving} proposes a new meta-learning regularization mechanism called Minimax-Meta Regularization, which uses inverse regularization in the inner loop and ordinary regularization in the outer loop during training. This continuously enhances the performance of the meta-learning algorithm.

In practical applications, meta-learning has increasingly been applied to semantic information research, with a primary focus on enhancing model robustness in complex and dynamic environments.  \cite{9189802
} proposed a meta-learning-based text-driven visual navigation model using semantic segmentation and Word2Vec to extract visual and semantic features, enabling quick adaptation to new tasks through meta-reinforcement learning. \cite{9174763} introduced Metabdry, a domain generalization method for named entity boundary detection, using meta-learning to simulate domain transfer and reduce performance differences. For few-shot semantic segmentation, \cite{PAMBALA202193} developed SML, which incorporates class-level semantic descriptions and ridge regression to compute prototypes, improving model performance with minimal data. Lastly, \cite{gong2021cluster} applied few-shot learning and meta-learning techniques to point cloud data semantic segmentation, effectively addressing issues related to label scarcity and class imbalance. Furthermore, by combining meta-learning with open composite domain adaptation, \cite{li2021fewshotmetalearningpointcloud} developed a new framework to address the open-label problem in semantic segmentation.

\subsection{Object Detection}
Modern object detection models are typically divided into two categories. \cite{zhang2021meta}. One set is region-based. The other is single-stage detectors, which do not require generating candidate regions. Instead, they transform the object detection task into a regression problem, significantly improving inference speed. Such as the YOLO framework \cite{redmon2016you} and SSD \cite{liu2016ssd}.

Currently, the single task of object detection is being increasingly integrated with other downstream tasks. In the context of low-light enhancement, \cite{wu2022edge} proposes an edge computing-driven, end-to-end framework for image enhancement and object detection under low-light conditions. This framework includes cloud-based enhancement and edge-based detection phases, significantly improving detection performance on edge devices while maintaining low latency in low-light environments. However, \cite{yuan2024plug} introduces a plug-and-play image enhancement model, also based on an end-to-end framework, which highlights the model’s ease of use and high efficiency in low-light conditions. Additionally, for the video analysis, \cite{zheng2022video} proposes a lightweight object detection network in fast-paced sports environments that reduces computational resource consumption and enhances detection robustness through multi-scale feature fusion and other techniques. Within this domain, object detection is further explored in target tracking applications. \cite{wang2025omnitracker} introduces a novel tracking-with-detection paradigm, where detection generates candidate bounding boxes in each video frame to improve the precision and robustness of target tracking. Recently, \cite{zhang2022bytetrack} introduced ByteTrack, which enhances multitarget tracking performance by extending the association of detection boxes, thereby preventing the loss of low-scoring targets. This approach further advances the field of multi-target tracking.

Object detection continues to see innovation in practical applications. \cite{alotaibi2022computational} introduces a new harmony search algorithm based on computational intelligence for real-time target detection and tracking in video surveillance systems, improving detection accuracy and tracking stability while efficiently identifying multiple targets in video frames. In autonomous vehicles fields, \cite{cai2021yolov4} proposes a single-stage object detection framework based on YOLOv4 to improve detection accuracy. Additionally, it introduces an optimized network pruning algorithm to address the challenge of meeting real-time performance requirements given the limited computational resources of vehicle-mounted platforms, thus improving the speed of model inference. In the field of digital medicine, object detection is also extensively applied in medical image analysis to assist doctors in accurately locating and identifying disease boundaries. \cite{gaikwad2024identification} combines YOLOv5 with deep neural networks to detect dislocations in vertebral columns. \cite{salman2022automated} uses the YOLO object detection algorithm to implement an automatic grading and diagnostic system for prostate cancer.

\section{Method}

The structure of SeFENet network is shown in Fig.~\ref{fig1:pipeine}, which contains 3 modules: Target Aware Homography Estimation Module (TAHEM), Semantic Extraction Module (SEM) and Semantic-guide Meta Constrains (SMC). First, we input the source image $I_\mathrm{s}$ and target image $I_\mathrm{t}$ into pre-trained TAHEM to get homography matrix. The hierarchical scale perception module is embedded in TAHEM to expand the receptive field of feature extraction, this approach enables the model to simultaneously capture both global and local information, allowing for a more comprehensive understanding of image content. Additionally, in challenging environments, it helps mitigate the impact of local noise, ensuring more robust and reliable results, so as to improve the robustness of the network in harsh environments. Then we input the ground truth $I_\mathrm{gt}$, the overlapping region of $I_\mathrm{s}$ and $I_\mathrm{t}$, into SEM to get semantic features. After that we feed this feature and the intermediate feature of TAHEM into SMC, implicitly enhance the ability of the proposed homography estimation network to The alternating optimization meta-learning strategy is employed to extract semantic features, enabling the model to capture multi-scale information while mitigating the impact of structural deformation in challenging environments. By leveraging semantic information to comprehend the contextual perspective relationships, the approach effectively reduces the likelihood of mismatches, thereby achieving robust homography estimation in degraded conditions. The specific details of key modules are as follows:

\subsection{Hierarchical Scale-aware}
In this module, we adopt a hierarchical attention mechanism that allows the network to focus on features at different scales across multiple iteration rounds. In the large-scale attention mechanism, we emphasize the global features of the image, capturing its overall structure. On the other hand, the small-scale attention mechanism concentrates on finer-grained detail features, enabling the network to capture subtle image information. By learning both global and local features simultaneously, our network improves its ability to perform homography estimation, leading to enhanced performance.

\begin{figure}[h]
    \centering
    \includegraphics[width=0.45\textwidth]{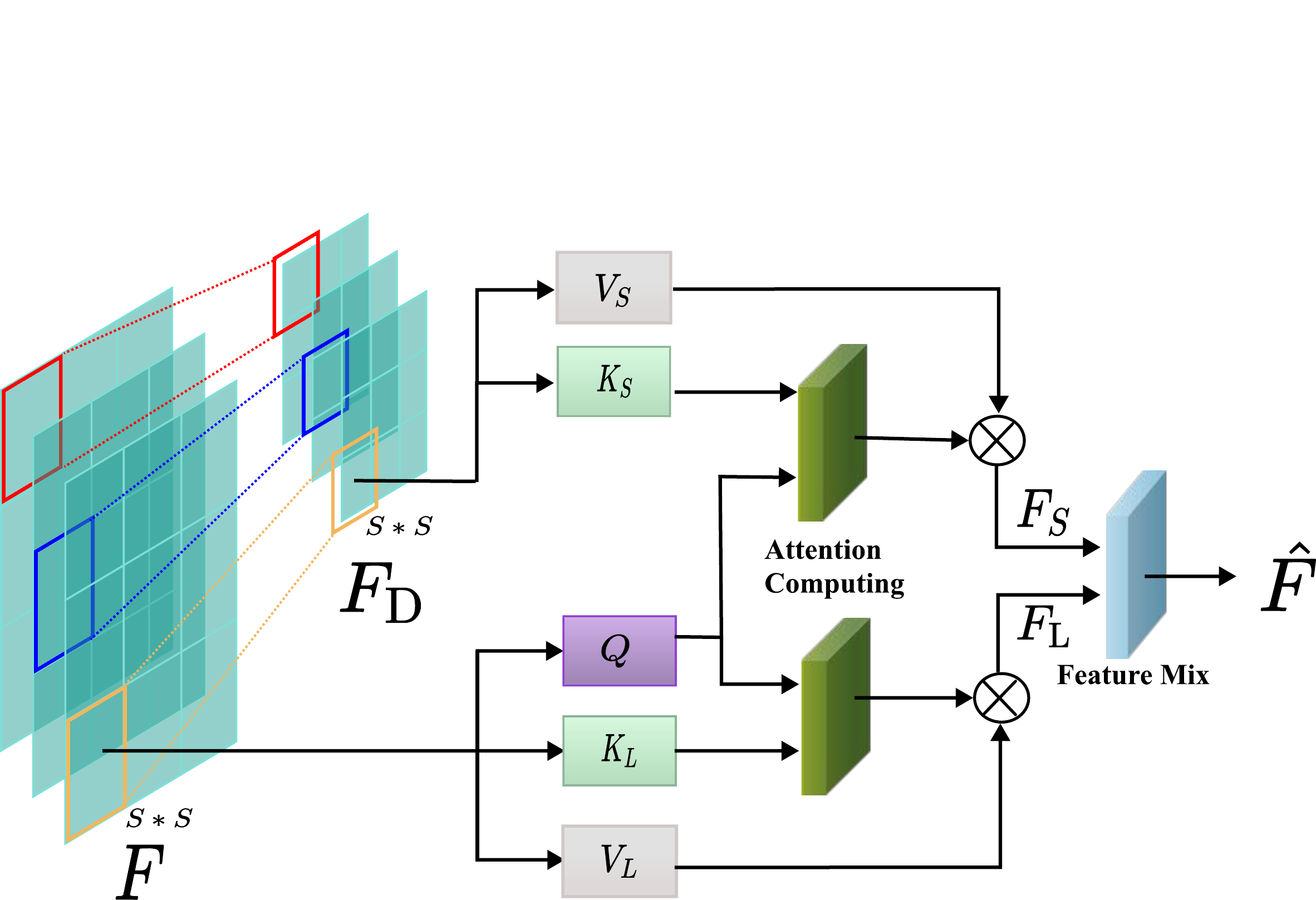}
    \caption{The implementation details of hierarchical scale-aware module. $F$ denotes the original feature. $F_\mathrm{D}$ denotes the feature after being downsampled. The pre-sampling is non-overlapping sampling and the sampling region will overlap after sampling. Same feature location is represented by same color. Multihead attention mechanism will be implemented interactively after sampling.}
    \label{fig1:cross}
\end{figure}
In harsh environments, traditional self-attention mechanism of the transformer has limited receptive field, which leads to insufficient image context information. To solve this problem, we adopt a hierarchical scale-aware mechanism to increase the receptive field during feature extraction. The details are shown in Fig.~\ref{fig1:cross}. We downsample the feature $F\in\mathbb{R}^{W*H*C}$ to $F_\mathrm{D} \in \mathbb{R}^{\frac{W}{2}*\frac{H}{2}*C}$. The size of hierarchical scale is $S*S$. We use it to chunk the feature before and after downsampling. The overlapping region of downsampled feature and hierarchical scale indicate the corresponding region, and the non-overlapping portion we fill in pads. Then we perform attention computation for the corresponding feature map. The computation process is shown below:
\begin{equation}
Q, K_\mathrm{L}, V_\mathrm{L} = \mathcal{F}(W(F)),
\label{equation:eq1}
\end{equation}
\begin{equation}
K_\mathrm{S}, V_\mathrm{S} = \mathcal{F}(W(F_\mathrm{D})),
\label{equation:eq2}
\end{equation}
where $W$ is the set of projection matrices and $\mathcal{F}$ is the function that computes the features $Q$, $K$ and $V$ under the 2 scales of features. We share the same $Q$ in 2 scales. $K_\mathrm{L}$ and $V_\mathrm{L}$ are the \textit{key} and \textit{value} in large scale. $K_\mathrm{S}$ and $V_\mathrm{S}$ are the \textit{key} and \textit{value} in small scale. The multi-head self-attention is then computed using the features before and after sampling separately in the \textit{Attention Computing} block as:
\begin{equation}
\text{Attention}(Q,K,V) = \text{softmax}\left(\frac{QK^T}{\sqrt{d}}\right) \cdot V,
\label{equation:eq3}
\end{equation}
where $Q$ is the shared \textit{query}. $K$ and $V$ are the \textit{key} and \textit{value} of the corresponding scale before and after sampling. $d$ is the \textit{query}/\textit{key} dimension. $F_\mathrm{L}$ and $F_\mathrm{S}$ are the results generated by the 2 scale self-attention mechanism respectively. Subsequently, these 2 results are fused through the convolutional layer and then connected in the network $\mathcal{M}$  to obtain the output feature $F_\mathrm{c}$, $F_\mathrm{c} = \mathcal{M}(F_\mathrm{L} + F_\mathrm{S})$. The feature boxes of the same color in the Fig.~\ref{fig1:cross} indicate the corresponding features before and after sampling. In this way, more representative information can be generated by this multi-scale feature extraction. In specific training, we use hierarchical scale-aware in \textit{Large scale Block} in Fig.~\ref{fig1:pipeine}(a), because the feature map with larger size contains more information and the information can be extracted more fully by hierarchical scale-aware. As for \textit{Small scale Block} in Fig.~\ref{fig1:pipeine}(a), the feature map has limited information because of the little size, thus we only use the normal self-attention strategy.
\begin{figure}[h]
    \centering
    \includegraphics[width=0.45\textwidth]{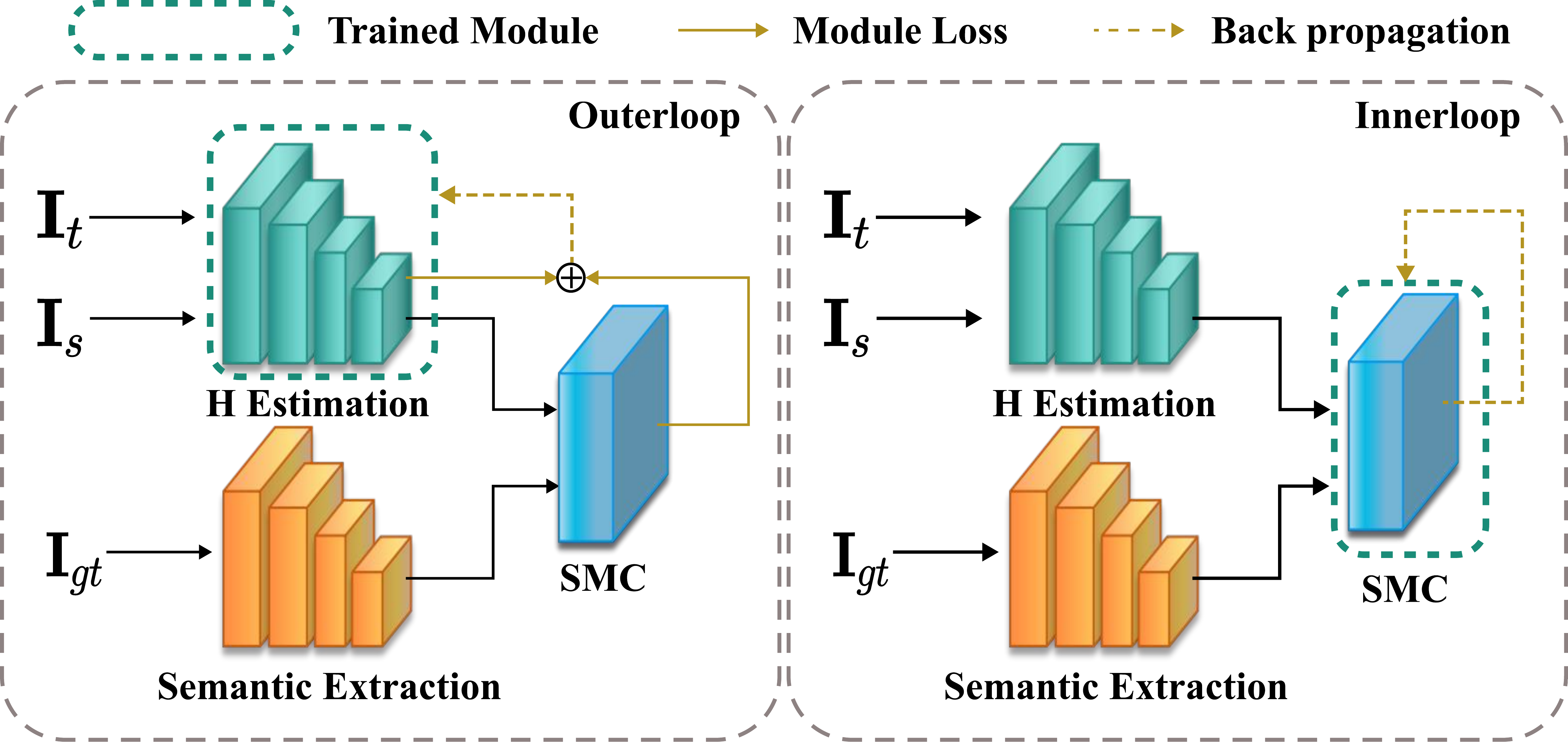}
    \caption{The details of the implementation of the meta-learning strategy are divided into 2 loops, inner and outer, in which the dashed box contains the module trained by backpropagation and the rest of the content is fixed.}
    \label{fig2:meta}
\end{figure}
\begin{figure*}[h]
    \centering
    \includegraphics[width=0.8\textwidth]{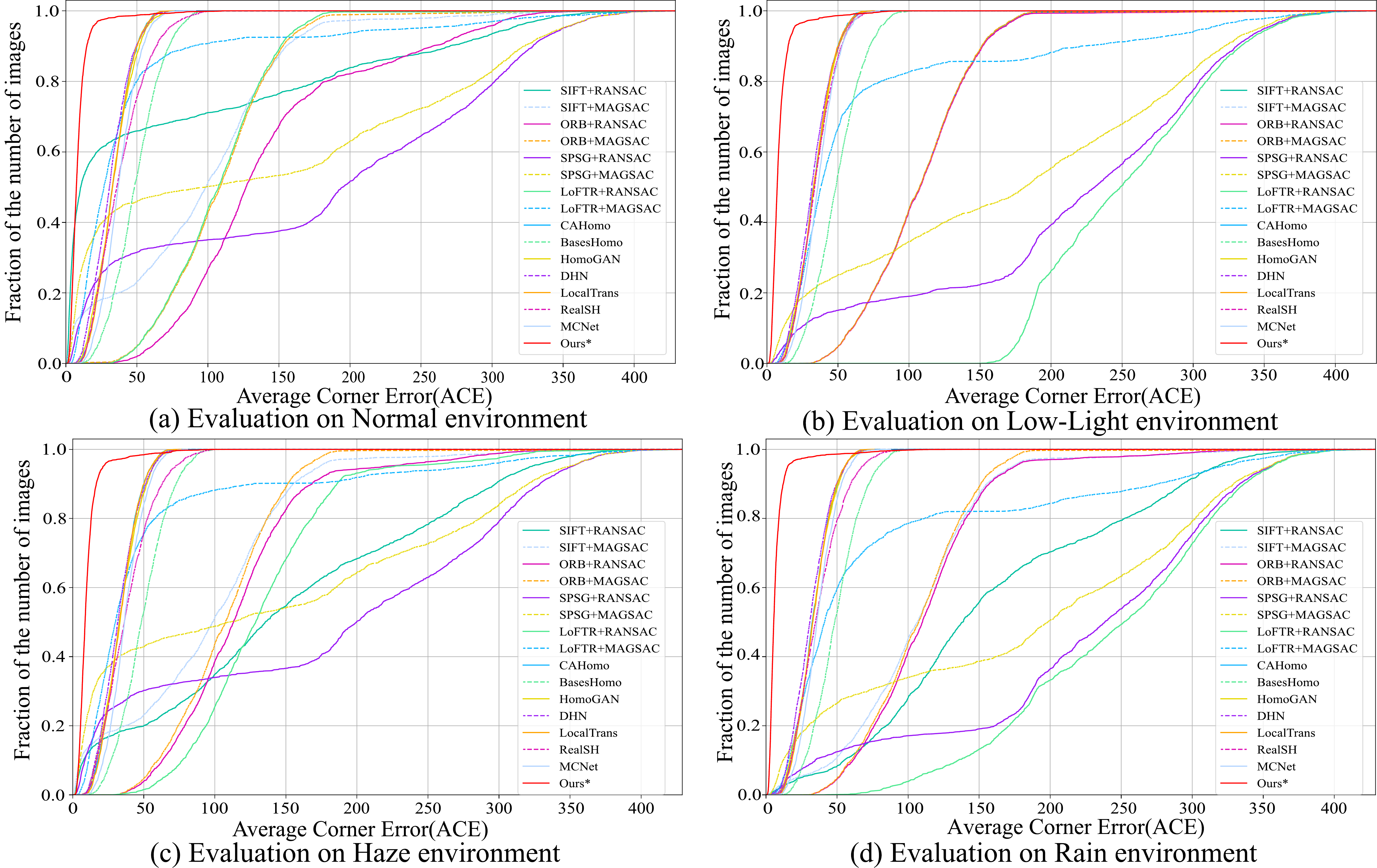}
    \caption{The 4 plots respectively indicate the evaluation of our homography estimation method on the VOC dataset in 4 environments: normal, low-light, haze, and rain. The x-axis represents the average corner error (ACE) and the y-axis represents the percentage of data under the corresponding ACE.}
    \label{fig3:voc_mace}
    \vspace{-10pt}
\end{figure*}

\subsection{Semantic-Guide Meta Constrains}
In this section, we integrate semantic features into the homography estimation network. While the existing homography networks primarily focus on the structural relationships between images, incorporating semantic features enables the network to also account for the semantic information at the model’s edges. This enhancement helps the network reduce issues such as ambiguity, leading to more accurate and robust homography estimation.

In order to fuse the semantic features extracted from the SEM and the structural features extracted from the TAHEM, we propose a Semantic-Guide Meta Constrains (SMC). The specific structure is shown in Fig.~\ref{fig1:pipeine}(c). Our SMC module has 2 components: Feature Mitigating Module (FMM) and Structure Transformation Module (STM). During the training process, we input $I_\mathrm{s}$ and $I_\mathrm{t}$ into TAHEM, and extract the structural features $F_\mathrm{t}^{1/8}$, $F_\mathrm{t}^{1/4}$ at 2 scales in \textit{Small Scale Block} and \textit{Large Scale Block}. Subsequently, the different scales of structural features are fed into STM separately. STM is a bridge to link the semantic features and structural features. It consists of 3 Conv + ReLU.

FMM can fuse structural and semantic features and then mitigate their differences in training. According to the size of features from TAHEM, we also extract the same size of semantic features containing $F_\mathrm{se}^{1/8}$, $F_\mathrm{se}^{1/4}$ from the SEM, which is a part of detection network named SSD~\cite{liu2016ssd}. Then we input the corresponding scale of semantic features and structural features into the FMM, the process as follows:
\begin{equation}
F_m = \mathcal{J}_\mathrm{5}( \mathcal{J}_\mathrm{4}(F_\mathrm{t}) \ \tikz[baseline=-0.75ex]\node[circle,draw,inner sep=1pt] {}; \ \mathcal{J}_\mathrm{2}(F_\mathrm{se})),
\label{equation:eq5}
\end{equation}
\begin{equation}
F = F_\mathrm{m} + \mathcal{J}_\mathrm{1}(F_\mathrm{m}),
\label{equation:eq6}
\end{equation}\begin{equation}
F_m = \mathcal{J}_\mathrm{5}( \mathcal{J}_\mathrm{4}(F_\mathrm{t}) \ \tikz[baseline=-0.75ex]\node[circle,draw,inner sep=1pt] {}; \ \mathcal{J}_\mathrm{2}(F_\mathrm{se})),
\label{equation:eq5}
\end{equation}
where $\mathcal{J}_\mathrm{n}$ denotes n layers of $3\times3$ Conv + ReLU and \ $\tikz[baseline=-0.75ex]\node[circle,draw,inner sep=1pt] {};$ \ denotes the concat operation between 2 features. From the features obtained by STM and FMM, we compute the loss constraining TAHEM and SMC networks, which will be explained in the next part. Our SEM network only assists the augmented homography estimation in the training phase and does not participate in the inference phase.

\begin{table*}[ht]
\centering
\small
\renewcommand{\arraystretch}{1.25}
\begin{tabular}{l|ccc|ccc|ccc|ccc}
\toprule
\multirow{2}{*}{Methods} &
  \multicolumn{3}{c|}{Normal} &
  \multicolumn{3}{c|}{Low-Light} &
  \multicolumn{3}{c|}{Haze} &
  \multicolumn{3}{c}{Rain} \\ 
  \cmidrule{2-13}
 &
  PSNR &
  SSIM &
  NCC &
  PSNR &
  SSIM &
  NCC &
  PSNR &
  SSIM &
  NCC &
  PSNR &
  SSIM &
  NCC \\ \midrule
1) SIFT+RANSAC &
  11.604 & 
  0.421 & 
  0.538 & 
  23.758 & 
  0.657 & 
  0.341 & 
  11.933 & 
  0.653 & 
  0.181 &
  11.507 & 
  0.398 & 
  0.487 \\
2) SIFT+MAGSAC &
  11.341 & 
  0.417 & 
  0.523 & 
  23.739 & 
  0.658 & 
  0.341 &
  10.934 &
  0.645 &
  0.124 &
  10.789 &
  0.425 &
  0.458 \\
3) ORB+RANSAC &
  11.719 &
  0.464 &
  0.478 &
  23.756 &
  0.659 &
  0.342 &
  11.453 &
  0.653 &
  0.148 &
  10.937 &
  \textcolor[RGB]{0,123,255}{\textbf{0.559}} &
  0.433 \\
4) ORB+MAGSAC &
  10.864 &
  0.548 &
  0.482 &
  23.740 &
  0.658 &
  0.341 &
  10.858 &
  0.646 &
  0.116 &
  10.344 &
  \textcolor[RGB]{204,0,0}{\textbf{0.628}} &
  0.424 \\
5) SPSG+RANSAC &
  11.357 &
  0.400 &
  0.523 &
  25.531 &
  0.629 &
  0.478 &
  13.215 &
  0.651 &
  0.489 &
  11.366 &
  0.336 &
  0.516 \\
6) SPSG+MAGSAC &
  11.326 &
  0.404 &
  0.529 &
  25.280 &
  0.652 &
  0.504 &
  12.694 &
  0.654 &
  0.465 &
  11.152 &
  0.329 &
  0.512 \\
7) LoFTR+RANSAC &
  11.110 &
  0.419 &
  0.501 &
  24.713 &
  0.641 &
  0.464 &
  12.203 &
  0.638 &
  0.386 &
  10.979 &
  0.335 &
  0.485 \\
8) LoFTR+MAGSAC &
  11.133 &
  0.419 &
  0.503 &
  24.793 &
  0.645 &
  0.473 &
  12.177 &
  0.640 &
  0.394 &
  10.979 &
  0.334 &
  0.487 \\ 
  \midrule
9) BasesHomo &
  19.028 &
  0.557 &
  0.750 &
  31.519 &
  0.856 &
  0.752 &
  34.760 &
  0.887 &
  0.831 &
  19.240 &
  0.395 &
  0.697 \\
10) HomoGAN &
  \textcolor[RGB]{204,0,0}{\textbf{20.496}} &
  \textcolor[RGB]{204,0,0}{\textbf{0.626}} &
  \textcolor[RGB]{0,123,255}{\textbf{0.842}} &
  \textcolor[RGB]{0,123,255}{\textbf{32.017}} &
  \textcolor[RGB]{0,123,255}{\textbf{0.869}} &
  0.880 &
  \textcolor[RGB]{0,123,255}{\textbf{35.859}} &
  \textcolor[RGB]{0,123,255}{\textbf{0.903}} &
  0.982 &
  \textcolor[RGB]{0,123,255}{\textbf{20.337}} &
  0.471 &
  0.830 \\ 
  \midrule
11) DHN & 
  18.651 &
  0.528 &
  0.722 &
  31.274 &
  0.846 &
  0.721 &
  34.932 &
  0.882 &
  0.810 &
  19.052 &
  0.378 &
  0.698 \\
12) LocalTrans &
  17.815 &
  0.480 &
  0.735 &
  31.192 &
  0.840 &
  0.807 &
  34.742 &
  0.881 &
  0.971 &
  18.671 &
  0.368 &
  0.739 \\
13) RealSH &
  14.415 &
  0.397 &
  0.498 &
  29.242 &
  0.810 &
  0.552 &
  31.737 &
  0.870 &
  0.865 &
  15.651 &
  0.304 &
  0.519 \\
14) MCNet &
  17.564 &
  0.577 &
  0.830 &
  31.602 &
  0.864 &
  \textcolor[RGB]{0,123,255}{\textbf{0.916}} &
  34.295 &
  0.898 &
  \textcolor[RGB]{204,0,0}{\textbf{0.992}} &
  18.774 &
  0.518 &
  \textcolor[RGB]{0,123,255}{\textbf{0.871}} \\
15) Ours &
  \textcolor[RGB]{0,123,255}{\textbf{20.445}} &
  \textcolor[RGB]{0,123,255}{\textbf{0.623}} &
  \textcolor[RGB]{204,0,0}{\textbf{0.892}} &
  \textcolor[RGB]{204,0,0}{\textbf{32.998}} &
  \textcolor[RGB]{204,0,0}{\textbf{0.887}} &
  \textcolor[RGB]{204,0,0}{\textbf{0.935}} &
  \textcolor[RGB]{204,0,0}{\textbf{36.612}} &
  \textcolor[RGB]{204,0,0}{\textbf{0.912}} &
  \textcolor[RGB]{0,123,255}{\textbf{0.991}} &
  \textcolor[RGB]{204,0,0}{\textbf{20.939}} &
  0.508 &
  \textcolor[RGB]{204,0,0}{\textbf{0.896}} \\ 
  
  \bottomrule  
\end{tabular}
\caption{The PSNR, SSIM, NCC of our method and other methods on large baseline VOC, where the best results are highlighted in red bold, and the second best results are highlighted in blue bold. For PSNR, SSIM, and NCC, a higher index value signifies better quality of the generated image.}
\label{table11:voc}
\end{table*}

\subsection{Meta-Learning strategy}
To better integrate semantic and structural information, we employ a meta-learning strategy. Through the interplay between the inner and outer loops, this strategy enables the model to optimize both the homography network and the feature fusion network simultaneously. As a result, the feature fusion network can more effectively assist in homography estimation, leading to improved performance.

Since the fixed SMC module parameters will affect the efficiency of combining semantic and structural features, we apply a meta-learning strategy to train the network, which can optimize both the TAHEM and the SMC modules so that the model can achieve better extraction and fusion ability. The specific training strategy for meta-learning is shown in Fig.~\ref{fig2:meta}. We divide the dataset into support dataset $D_s$ and query dataset $D_\mathrm{q}$, where the query dataset is subdivided into training set $D_\mathrm{qr}$ and test set $D_\mathrm{qt}$.

Our training process contains 2 parts: innerloop and outerloop. For the innerloop, we only optimize the SMC module on the $D_\mathrm{qr}$, while both TAHEM module and SEM module are in a fixed state. The formula of the process of updating SMC parameters is as follows:
\begin{align}
    \theta_\mathrm{SMC} &= \theta_\mathrm{SMC} - \eta_\mathrm{SMC} \nabla_{\theta_\mathrm{SMC}} \big( \mathcal{L}_\mathrm{m}(F_\mathrm{se}^{1/8},F_{t}^{1/8}) \notag \\
    &\quad + \mathcal{L}_\mathrm{m}(F_\mathrm{se}^{1/4},F_\mathrm{t}^{1/4}) \big),
    \label{equation:eq7}
\end{align}
where $\theta_\mathrm{SMC}$ denotes the parameters before and after the SMC update. $\eta_\mathrm{SMC}$ is the learning rate of the SMC module. $\nabla_{\theta_\mathrm{SMC}} \mathcal{L}_\mathrm{m}(F_\mathrm{se},F_\mathrm{t})$ can be calculated by:
\begin{align}
    \nabla_{\theta_\mathrm{SMC}} \mathcal{L}_\mathrm{m}(F_\mathrm{se},F_\mathrm{t}) = \frac{\partial \mathcal{L}_\mathrm{m}(F_\mathrm{se},F_\mathrm{t})}{\partial \theta_\mathrm{SMC}},
\label{equation:eq8}
\end{align}
$\mathcal{L}_\mathrm{m}$ is the loss function of the SMC module, matching $L_2$ distance. We use it to improve the ability of the SMC to mitigate the difference between semantic and structural features.
In the outerloop, we fix the SMC and SEM modules to train the TAHEM by $D_s$. For our TAHEM network, we use the loss of ~\cite{cao2023recurrent} to compute the $L_1$ distance between the prediction result and the ground truth in each iteration, named $\mathcal{L}_\mathrm{h}$. In the training, we use the combination of $\mathcal{L}_\mathrm{h}$ and $\mathcal{L}_\mathrm{m}$ losses to update the module, the process as follows:
\begin{align}
    \theta_\mathrm{TAHEM} &= \theta_\mathrm{TAHEM} - \eta_\mathrm{TAHEM} \nabla_{\theta_\mathrm{TAHEM}} \big( \mathcal{L}_\mathrm{h}(I_{s},I_{t},I_{gt}) \notag \\
    &\quad + \lambda \sum \mathcal{L}_\mathrm{m}(F_\mathrm{se},F_\mathrm{t}) \big).
\label{equation:eq9}
\end{align}
Similarly, the derivation of the loss is done as follows:
\begin{align}
    &\nabla_{\theta_\mathrm{TAHEM}} \big(\mathcal{L}_\mathrm{h}(I_{s},I_{t},I_{gt}) + \lambda \sum \mathcal{L}_\mathrm{m}(F_\mathrm{se},F_\mathrm{t})\big) \notag \\
    &= \frac{\partial \mathcal{L}_\mathrm{h}(I_\mathrm{s},I_\mathrm{t},I_\mathrm{gt})}{\partial \theta_\mathrm{TAHEM}} 
    +  \lambda \sum \frac{\partial \mathcal{L}_\mathrm{m}(F_\mathrm{se},F_\mathrm{t})}{\partial \theta_\mathrm{TAHEM}}, 
\label{equation:eq10}
\end{align}
where $I_\mathrm{s}$ denotes the source image. $I_\mathrm{t}$ denotes the target image. $I_\mathrm{gt}$ denotes the ground truth. $\theta_\mathrm{TAHEM}$ is the parameters of the TAHEM module before and after the update and $\eta_\mathrm{TAHEM}$ is the learning rate of the TAHEM. $\lambda$ is a hyper-parameter used to balance $\mathcal{L}_\mathrm{h}$ and $\mathcal{L}_\mathrm{m}$. $\sum \mathcal{L}_\mathrm{m}(F_{se},F_{t})$ denotes the sum of the losses at the size of $\frac{1}{4}$ and $\frac{1}{8}$.

During the training process, we first pre-train a SEM and TAHEM, then we train pre-trained model by meta-learning. In addition to adding hyper-parameter $\lambda$, to balance $\mathcal{L}_\mathrm{h}$ and $\mathcal{L}_\mathrm{m}$ , we optimize SMC once while optimizing TAHEM 3 times. After fixed epochs, we use $D_\mathrm{qt}$ to test the model.

\section{Experiment}
\subsection{Implementation Details}
In our experiments, we set the hyper-parameter $\lambda$ to 0.1, the number of iteration per scale $n$ to 5. In the hierarchical scale-aware module, we set the size of the sampled hierarchical scale $S$ to 4. Our network is done with Pytorch, using NVIDIA RTX 4090Ti GPU. During training, we use AdamW optimizer for TAHEM and SMC, with the learning rate set to \num{1e-4}, pre-training with 20 epochs at a batchsize of 32, training with 50 epochs at a batchsize of 16. We also use SGD optimizer for SEM, with the leaning rate set to \num{1e-3}, training with 20 epochs at batchsize of 32.
\subsection{Dataset and Experiment Settings}
Considering the need to incorporate semantic conditional information during training, we first create a synthetic homography estimation dataset using the VOC detection dataset ~\cite{everingham2010pascal}. To assess the robustness of our model in real-world environments, we utilize the large-baseline CAHomo dataset proposed in ~\cite{jiang2023semi}. In order to simulate harsh conditions typical of real-world scenarios, we add elements such as rain, haze, and low-light to both the large-baseline CAHomo and VOC datasets.

\begin{figure*}[h]
    \centering
    \includegraphics[width=0.9\textwidth]{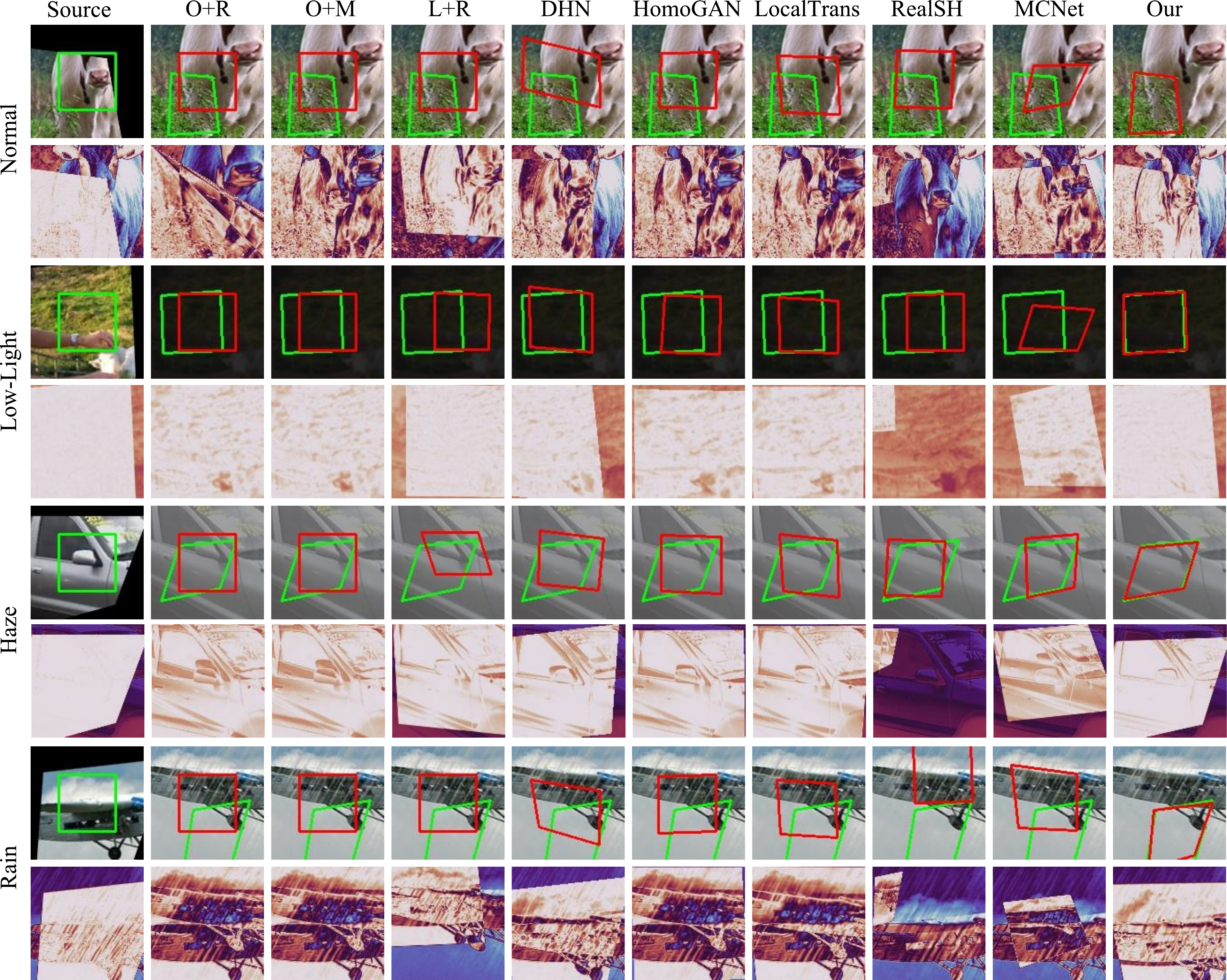}
    \caption{The first row in each environment displays visualization results of some traditional and deep learning methods on the VOC dataset, where the green polygon indicates the position of the source image in terms of the truth value on the target image. The red box indicates the prediction in the target image after different homography estimation algorithms. The closer the distance between the 2 polygonal boxes, the better the homography estimation.
 The second row in each environment shows the error image, which is the difference between the warped image and the target image. The closer the error image is to the ground truth, indicating more accurate homography estimation. Here, O+R refers to ORB+RANSAC, O+M to ORB+MAGSAC, and L+R to LoFTR+RANSAC.}
    \label{fig3:voc_visual}
\end{figure*}
\begin{figure*}[h]
    \centering
    \includegraphics[width=1\textwidth]{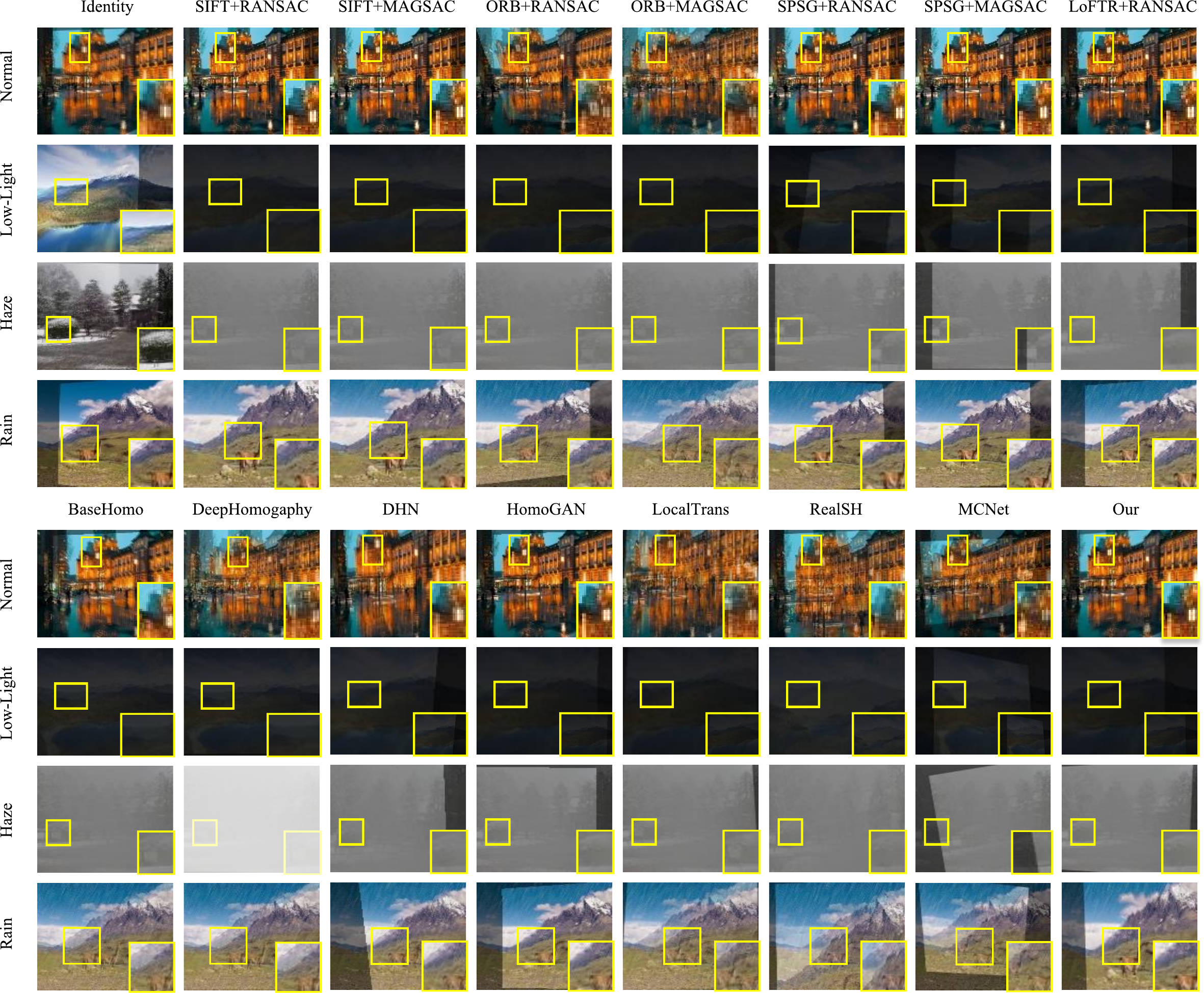}
    \caption{Qualitative analysis with other methods on large-baseline CAHomo, the yellow box is the part that is prone to error during homography transformation and the yellow box in the lower right corner is the effect of viewing it enlarged.}
    \label{fig4:CA_visual}
\end{figure*}

\textbf{Experiment Settings}: As in previous works ~\cite{detone2016deep,erlik2017homography,le2020deep}, we set the input image size to [128×128], then randomly perturb and offset the 4 endpoints. To generate the synthetic dataset, we set the offset overlap to 0.5, with the range of random perturbation set to [-25.6, 25.6]. For the VOC dataset, we use mean average corner error (MACE) as our evaluation metric. MACE calculates the mean of the mean squared error (MSE) between the ground truth and the estimated positions of the 4 corner points. For the large-baseline CAHomo dataset, we use point matching errors (PME) as evaluation metric, representing the mean difference between the feature points after homography estimation and the ground truth. Additionally, other metrics are also used such as Peak Signal-to-Noise Ratio (PSNR), Structural Similarity Index (SSIM) ~\cite{wang2004image} and Normalized Cross-Correlation (NCC) ~\cite{zitova2003image} to evaluate the quality of the generated images after homography estimation.

We evaluate our model on comparison experiments including 3 kinds of existing different homography estimation methods: 1)Traditional methods: the feature point extraction methods include SIFT~\cite{lowe2004distinctive}, ORB~\cite{rublee2011orb}, SuperPoint~\cite{detone2018superpoint} with SuperGlue~\cite{sarlin2020superglue} (SPSG) and LoFTR~\cite{sarlin2020superglue}. Outlier culling based on the traditional methods include RANSAC~\cite{fischler1981random} and MAGSAC~\cite{barath2019magsac}. We combine these 4 fetching point methods and 2 outlier point methods to get comparison methods. 2)Unsupervised methods: including CAHomo~\cite{zhang2020content}, BasesHomo~\cite{ye2021motion} and HomoGAN~\cite{hong2022unsupervised}. 3)Supervised methods: including DHN~\cite{detone2016deep}, LocalTrans~\cite{shao2021localtrans}, RealSH~\cite{jiang2023supervised} and MCNet~\cite{zhu2024mcnet}.

\begin{table*}[ht]
\centering
\small
\setlength{\tabcolsep}{0.55mm}
\renewcommand{\arraystretch}{1.25}
\begin{tabular}{l|cccc|cccc|cccc|cccc}
\toprule
\multirow{2}{*}{Methods} &
  \multicolumn{4}{c|}{Normal} &
  \multicolumn{4}{c|}{Low-Light} &
  \multicolumn{4}{c|}{Haze} &
  \multicolumn{4}{c}{Rain} \\ 
  \cmidrule{2-17}
 &
  PME &
  PSNR &
  SSIM &
  NCC &
  PME &
  PSNR &
  SSIM &
  NCC &
  PME &
  PSNR &
  SSIM &
  NCC &
  PME &
  PSNR &
  SSIM &
  NCC \\ \midrule
1) SIFT+RANSAC &
  19.669 &
  11.604 &
  0.420 &
  0.537 &
  69.175 &
  23.754 &
  0.656 &
  0.341 &
  68.755 &
  11.932 &
  0.652 &
  0.180 &
  32.341 &
  11.507 &
  0.397 &
  0.486 \\
2) SIFT+MAGSAC &
  18.543 &
  11.341 &
  0.417 &
  0.522 &
  69.175 &
  23.739 &
  0.658 &
  0.341 &
  68.831 &
  10.934 &
  0.644 &
  0.123 &
  31.126 &
  10.788 &
  0.424 &
  0.457 \\
3) ORB+RANSAC &
  53.023 &
  11.718 &
  0.463 &
  0.477 &
  69.253 &
  23.755 &
  0.658 &
  0.342 &
  69.253 &
  11.453 &
  0.652 &
  0.148 &
  67.544 &
  10.937 &
  \textcolor[RGB]{0,123,255}{\textbf{0.558}} &
  0.433 \\

4) ORB+MAGSAC &
  50.790 &
  10.774 &
  0.532 &
  0.471 &
  69.253 &
  23.740 &
  0.658 &
  0.341 &
  69.253 &
  10.857 &
  0.645 &
  0.116 &
  66.806 &
  10.344 &
  \textcolor[RGB]{204,0,0}{\textbf{0.628}} &
  0.423 \\
5) SPSG+RANSAC &
  21.353 &
  11.357 &
  0.400 &
  0.522 &
  37.697 &
  25.531 &
  0.628 &
  0.478 &
  35.562 &
  13.215 &
  0.650 &
  0.489 &
  26.235 &
  11.365 &
  0.335 &
  0.516 \\
6) SPSG+MAGSAC &
  14.454 &
  11.325 &
  0.404 &
  0.528 &
  28.784 &
  25.279 &
  0.651 &
  0.504 &
  22.927 &
  12.694 &
  0.653 &
  0.465 &
  18.349 &
  11.152 &
  0.328 &
  0.512 \\
7) LoFTR+RANSAC &
  17.449 &
  11.109 &
  0.419 &
  0.500 &
  15.571 &
  24.712 &
  0.640 &
  0.463 &
  24.637 &
  12.202 &
  0.637 &
  0.386 &
  18.459 &
  10.978 &
  0.335 &
  0.485 \\
8) LoFTR+MAGSAC &
  16.085 &
  11.132 &
  0.418 &
  0.503 &
  15.731 &
  24.792 &
  0.644 &
  0.473 &
  24.207 &
  12.177 &
  0.639 &
  0.393 &
  16.818 &
  10.979 &
  0.333 &
  0.486 \\ \midrule
9) CAHomo &
  76.271 &
  8.938 &
  0.253 &
  0.269 &
  71.546 &
  21.952 &
  0.465 &
  0.193 &
  71.392 &
  9.719 &
  0.473 &
  0.096 &
  75.359 &
  8.663 &
  0.189 &
  0.227 \\
10) BasesHomo &
  15.567 &
  19.028 &
  0.557 &
  0.749 &
  14.467 &
  31.518 &
  0.855 &
  0.751 &
  13.815 &
  34.760 &
  0.886 &
  0.830 &
  14.441 &
  19.240 &
  0.395 &
  0.697 \\
11) HomoGAN &
  18.394 &
  \textcolor[RGB]{0,123,255}{\textbf{20.495}} &
  \textcolor[RGB]{0,123,255}{\textbf{0.625}} &
  \textcolor[RGB]{0,123,255}{\textbf{0.841}} &
  15.870 &
  \textcolor[RGB]{0,123,255}{\textbf{32.017}} &
  \textcolor[RGB]{0,123,255}{\textbf{0.869}} &
  0.880 &
  15.220 &
  \textcolor[RGB]{0,123,255}{\textbf{35.858}} &
  \textcolor[RGB]{0,123,255}{\textbf{0.902}} &
  0.981 &
  17.591 &
  \textcolor[RGB]{0,123,255}{\textbf{20.337}} &
  0.470 &
  \textcolor[RGB]{0,123,255}{\textbf{0.830}} \\ 
  \midrule
12) DHN &
  63.891 &
  18.650 &
  0.528 &
  0.722 &
  67.690 &
  31.273 &
  0.846 &
  0.721 &
  67.437 &
  34.932 &
  0.881 &
  0.809 &
  64.388 &
  19.051 &
  0.377 &
  0.698 \\
13) LocalTrans &
  12.631 &
  17.814 &
  0.480 &
  0.734 &
  12.801 &
  31.191 &
  0.839 &
  0.806 &
  12.705 &
  34.742 &
  0.881 &
  0.970 &
  12.693 &
  18.670 &
  0.367 &
  0.738 \\
14) RealSH &
  27.287 &
  14.414 &
  0.397 &
  0.497 &
  42.284 &
  29.242 &
  0.810 &
  0.552 &
  45.465 &
  31.737 &
  0.870 &
  0.864 &
  33.601 &
  15.651 &
  0.303 &
  0.518 \\
15) MCNet &
  \textcolor[RGB]{0,123,255}{\textbf{11.658}} &
  17.738 &
  0.505 &
  0.795 &
  \textcolor[RGB]{0,123,255}{\textbf{11.943}} &
  31.442 &
  0.851 &
  \textcolor[RGB]{0,123,255}{\textbf{0.883}} &
  \textcolor[RGB]{0,123,255}{\textbf{12.704}} &
  34.732 &
  0.893 &
  \textcolor[RGB]{0,123,255}{\textbf{0.987}} &
  \textcolor[RGB]{0,123,255}{\textbf{12.053}} &
  18.868 &
  0.418 &
  0.825 \\
16) Ours &
  \textcolor[RGB]{204,0,0}{\textbf{5.186}} &
  \textcolor[RGB]{204,0,0}{\textbf{20.821}} &
  \textcolor[RGB]{204,0,0}{\textbf{0.632}} &
  \textcolor[RGB]{204,0,0}{\textbf{0.891}} &
  \textcolor[RGB]{204,0,0}{\textbf{5.394}} &
  \textcolor[RGB]{204,0,0}{\textbf{32.995}} &
  \textcolor[RGB]{204,0,0}{\textbf{0.886}} &
  \textcolor[RGB]{204,0,0}{\textbf{0.934}} &
  \textcolor[RGB]{204,0,0}{\textbf{7.043}} &
  \textcolor[RGB]{204,0,0}{\textbf{36.612}} &
  \textcolor[RGB]{204,0,0}{\textbf{0.912}} &
  \textcolor[RGB]{204,0,0}{\textbf{0.990}} &
  \textcolor[RGB]{204,0,0}{\textbf{6.073}} &
  \textcolor[RGB]{204,0,0}{\textbf{20.939}} &
  0.507 &
  \textcolor[RGB]{204,0,0}{\textbf{0.895}} \\ 
  
  \bottomrule  
\end{tabular}
\caption{The point matching errors(PME), PSNR, SSIM, NCC of our method and other methods on large baseline CAHomo, where the best results are highlighted in red bold, and the second best results are highlighted in blue bold. For PME, a lower index value indicates better homography estimation. For PSNR, SSIM, and NCC, a higher index value signifies better quality of the generated image.}
\label{table1:pme}
\end{table*}

\subsection{Evaluation on VOC dataset}
We conducted a comprehensive evaluation on the improved VOC benchmark dataset in normal conditions and 3 harsh environments. To assess the quality of the generated images, we utilized PSNR, SSIM, and NCC, as presented in Table.~\ref{table11:voc}. As shown in the table, our approach exhibits clear advantages. It consistently ranked among the top two in nearly all test environments and achieved the best performance across most metrics. In the Normal environment, although PSNR and SSIM are slightly lower than those of HomoGAN, the average difference is only 0.365\%, making the gap negligible. In contrast, the NCC is significantly higher, surpassing HomoGAN by nearly 6\%, demonstrating stronger feature consistency. In complex environments, with the exception of the NCC in the Haze environment and the SSIM in the Rain environment, most other metrics outperform all competing methods, achieving an average improvement of nearly 3\% compared to the second-best approach. Notably, in Low-Light environments, our method exhibited absolute superiority, outperforming all other methods across all evaluation metrics, further underscoring its exceptional capability in feature point extraction under harsh conditions. Overall, our approach excels in preserving structural information and visual quality, demonstrating greater adaptability and robustness in both standard and challenging environments. 

Then, in Fig.~\ref{fig3:voc_mace}, we compare the performance of 16 methods on the VOC dataset in 4 environments. The results show that harsh environments make feature point extraction challenging, leading the convergence speed of the proportion of different ace images to 1 is much slower than the method based on deep learning. However, with the growth of ACE, our method can make the proportion of images with different ACE converging to 1 fastest.

We visualized the homography estimation results in the first row of each environment in Fig.~\ref{fig3:voc_visual}. Traditional methods such as ORB+RANSAC, ORB+MAGSAC, and LoFTR+RANSAC exhibit significant instability, failing to produce satisfactory results across all 4 scenarios. For other deep learning methods, the synthetic large baseline dataset also poses a challenge due to increased noise in harsh environments, resulting in a considerable deviation between their results and the ground truth. In contrast, our method shows a nearly perfect match between the red box and the green box,

\begin{figure}
    \centering
    \includegraphics[width=0.485\textwidth]{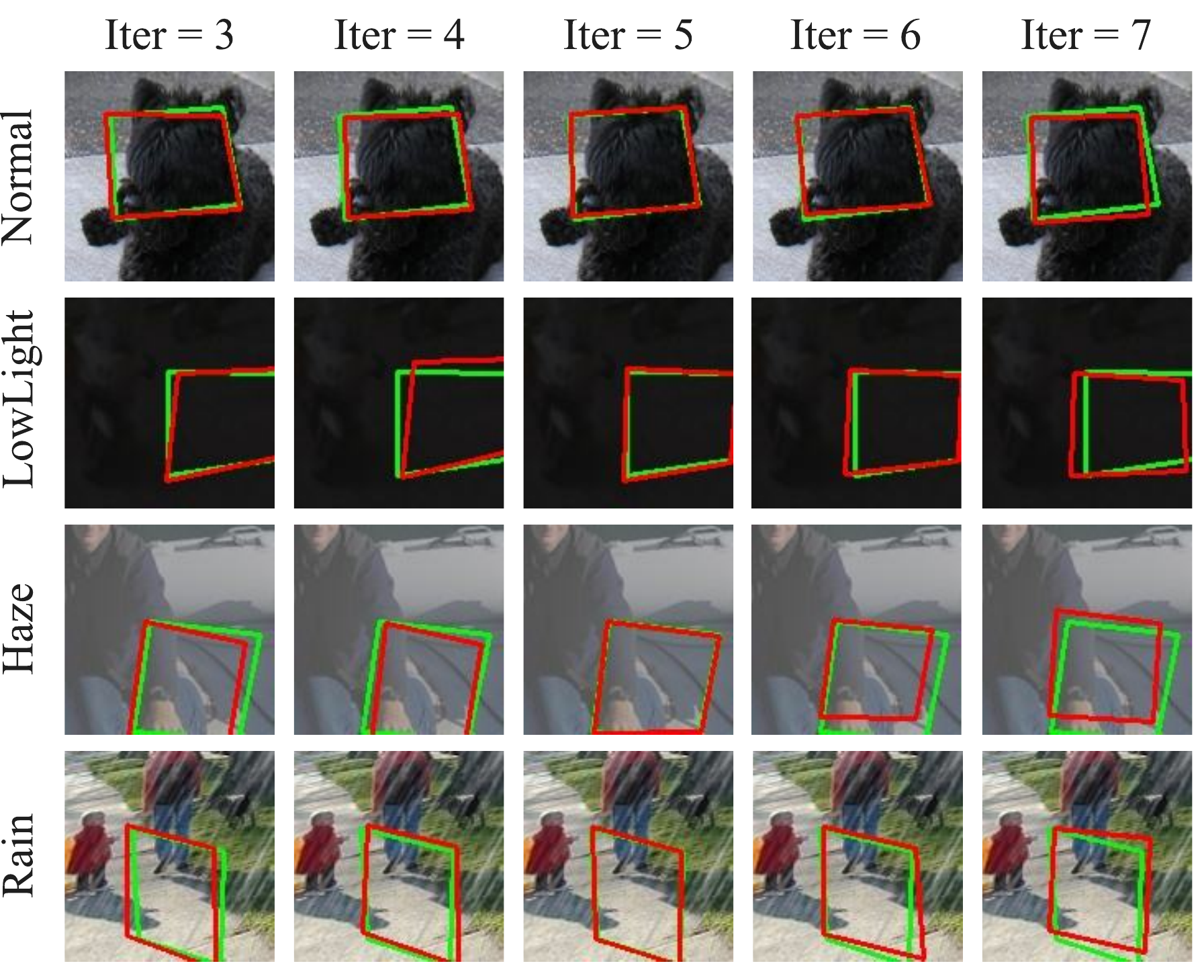}
    \caption{The ablation of different iterations, where the green polygon indicates the position of the source image in terms of the truth value on the target image. The red box indicates the prediction in the target image.Iter = 5 is our method.}
    \label{fig:iter}
\end{figure}
Meanwhile, we visualized the error between the warped image and the target image, as shown in the second row in Fig.~\ref{fig3:voc_visual}. Under the normal environment, although the overall error is small, some methods, such as O+M, DHN, and HomoGAN, exhibit large errors when compared to the error map generated by the true values, especially in the areas around the target box. The predictions of these methods deviate from the position of the source image, resulting in misalignment errors. In contrast, the error map generated by our method is almost perfectly consistent with the true value. In harsh environments, due to feature occlusion and blurring caused by environmental noise, most methods show large errors. Notably, in rainy conditions, RealSH and MCNet exhibit the most significant errors, with serious image misalignment that almost completely deviates from the true values. However, in these complex environments, our method performs particularly well. Even when affected by noise, the resulting error maps are smaller and more consistent with the true value compared to those generated by other methods. In summary, compared to other methods, our model get better homography matrix various environments. There is no obvious misalignment in the error maps, and they remain highly consistent with the error maps generated by the true values, further confirming the accuracy of our method in homography estimation.

\begin{figure*}[h]
    \centering
    \includegraphics[width=1\textwidth]{heapmap.pdf}
    \caption{The features extracted from models with and without meta-learning strategies are visualized using heap maps in normal, low-light, haze and rain environments. Cool color areas indicate areas where features are of higher importance. Warm color areas represent areas with less feature information. ML means meta-learning.}
    \label{fig:heapmap}
    \vspace{-10pt}
\end{figure*}
\begin{figure}
    \centering
    \includegraphics[width=0.485\textwidth]{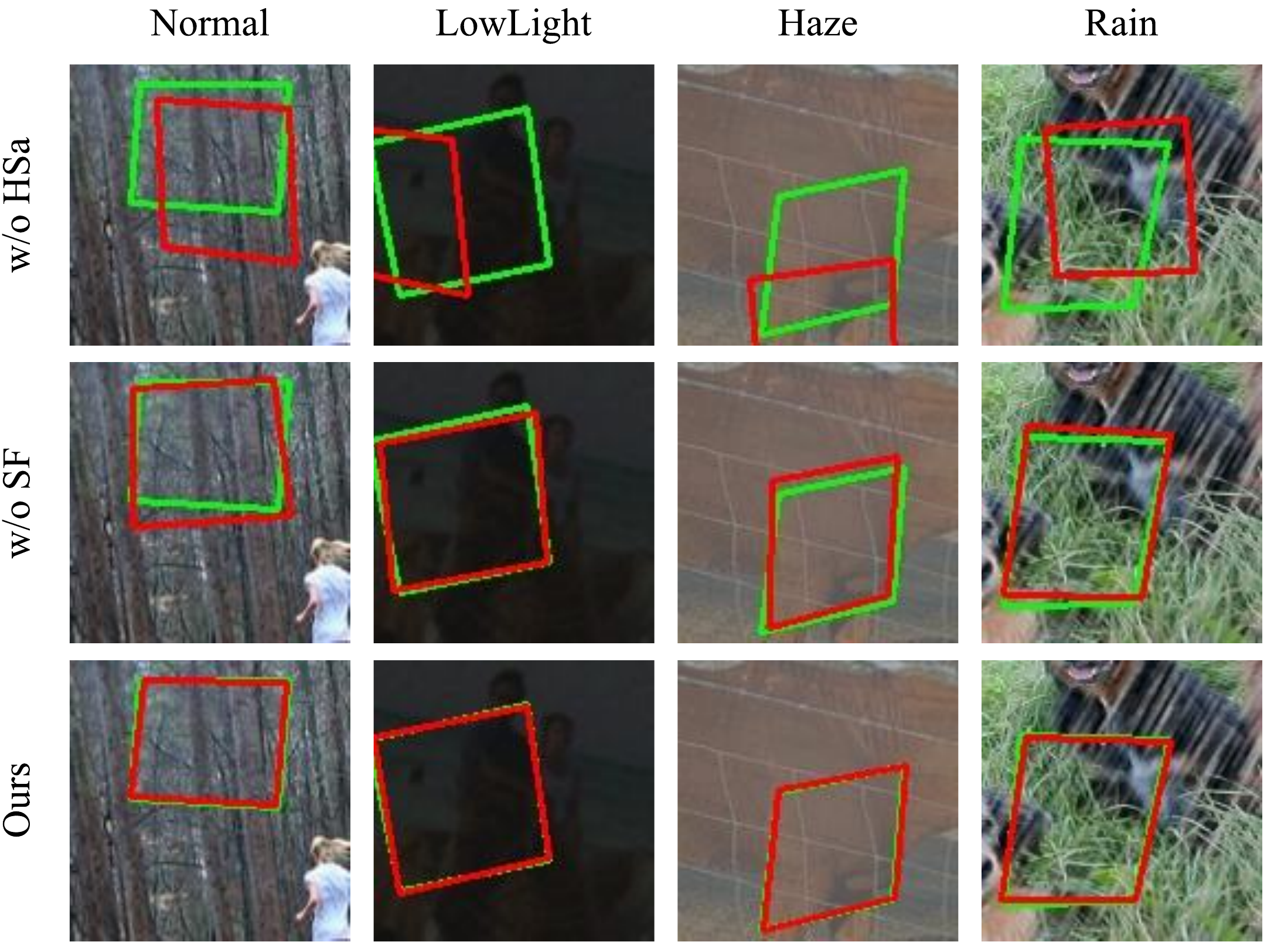}
    \caption{The ablation of our network, from top to bottom are no Hierarchical Scale-aware network, no semantic guide module and our model. HSa represents Hierarchical Scale-aware network and SF means Semantic Feature guide module.}
    \label{fig:tran}
    \vspace{-10pt}
\end{figure}

\subsection{Evaluation on Large-Baseline CAHomo dataset}
To demonstrate the robustness of our method in real environments, we test it using an improved benchmark dataset based on CAHomo from ~\cite{jiang2023semi}. We employ the same comparison method as the VOC dataset test. PME is used to evaluate the homography estimation results, while PSNR, SSIM, and NCC assess the quality of the generated images. The experimental results are presented in Table.~\ref{table1:pme}, showing that traditional feature point extraction methods are often unstable, leading to awful outcomes. Except for the LoFTR-based point fetching method, PME for images generated by traditional methods is much lower in natural conditions compared to harsh environments. Most deep learning methods exhibit better robustness with images having large baselines. However, our method shows a clear advantage over others in both normal and harsh environments. Compared to the PME of the second-place MCNet, our model reduces PME by 55.5\%, 54.8\%, 41.7\%, and 49.6\% across the 4 cases respectively. Additionally, our method performs well in PSNR, SSIM, and NCC. Traditional image generation methods often produce subpar results due to the challenges associated with feature point extraction in harsh environments, resulting in equal values of PSNR, SSIM, and NCC for some methods under certain conditions. While our method's SSIM only lags behind ORB+MAGSAC and ORB+RANSAC in the rain case, it ranks first in all other metrics across all conditions, with at least a 3\% improvement over the second-place metrics.

indicating highly accurate homography estimation. 
\begin{figure*}[h]
\includegraphics[width=0.995\textwidth]{Iters.pdf}
    \caption{Metrics—including MACE, PSNR, SSIM, and NCC—were computed from both the VOC dataset and the CAHomo dataset and evaluated across iterative blocks (3 to 7) under normal, low-light, haze, and rain conditions. Lower MACE and PME values indicate better performance, while higher PSNR, SSIM, and NCC values reflect improvement.}
    \label{fig5:albation}
\end{figure*}
\begin{figure*}[h]
\includegraphics[width=0.995\textwidth]{Three.pdf}
    \caption{The metrics calculated from the VOC dataset and CAHomo dataset remain consistent with previous evaluations but are used here to compare the performance of our model with and without the fusion of semantic features or hierarchical scale-awareness. Lower MACE and PME values indicate better performance, while higher PSNR, SSIM, and NCC values reflect improvement.}
    \label{fig7:albation}
    \vspace{-10pt}
\end{figure*}

In Fig.~\ref{fig4:CA_visual}, we visualize the results of homography estimation on the large-baseline CAHomo dataset across 4 different environments for each method. From Fig.~\ref{fig4:CA_visual}, it is evident that deep learning-based methods exhibit greater robustness compared to traditional feature point-based approaches, except for LoFTR+RANSAC. However, in harsh environments, some cases still exhibit heavy shadows in critical areas or significant distortion deviations. In contrast, our method effectively mitigates shadows and distortion at the edges, resulting in distorted target images that closely resemble the source images.

\subsection{Ablation Study}
\textbf{Optimal number of iteration modules}: Our model includes an iteration module that recurrently extracts and estimates features at each scale. To determine the optimal number of iteration modules, we experimented with \( n = 3, 4, 5, 6, \) and \( 7 \). The similar results in Fig.~\ref{fig5:albation} illustrate how the performance evolves as the number of iterations changes, providing a clearer understanding of the model’s behavior under different configurations. The first row of Fig.~\ref{fig5:albation} represents the VOC dataset, and the second row represents the CAHomo dataset. From Fig.~\ref{fig5:albation} (a), it is clear that when the number of iterations is set to 5, both the MACE values and PME values reach their lowest points, indicating optimal performance. When the number of iterations exceeds 5, the MACE values and PME values increase, resulting in performance deterioration. For the PSNR, SSIM, and NCC indicators across the two datasets, although the trend is not as pronounced as the above two indicators, in most cases, these values all remain at a high level when iteration is 5, which is better than when iteration is increased. This further confirms the superior performance of the five-iteration model. From the perspective of the model network architecture, this suggests that as the number of iterations increases, the network can begin to focus more on irrelevant or noisy details, rather than the meaningful features required for accurate homography estimation. Furthermore, from a model generalization perspective, another key factor to consider is the computational cost associated with increasing the number of iterations. As iterations increase, training time grows, which demands more computational resources, but also raises the possibility of inefficiency. In summary, considering the balance between model performance and training cost, we set the number of iterations in the model to 5.

The results of ablation experiments with different iteration blocks are shown in Fig.~\ref{fig:iter}, which shows the results visualized with different iteration blocks using the pre-trained model. As shown in the figure, our matching effect is the best when iteration block=5. As the number of blocks increases or decreases, the gap between the prediction box and the ground truth box gradually increases.

\textbf{Effectiveness of hierarchical scale-aware and semantic extraction}: To demonstrate the effectiveness of hierarchical scale-aware and semantic features, we conducted ablation studies by removing these 2 modules separately from the model and then training it without them on two datasets. To demonstrate the effectiveness of the hierarchical scale-aware and semantic feature modules, we conducted ablation studies by separately removing these two modules from the model and training it on two datasets without them. In the experiment testing the impact of the hierarchical scale-aware module, we used features extracted by the model without this module for correlation computation. As shown in Fig.~\ref{fig7:albation}, the values of the five indicators significantly deteriorate when the hierarchical scale-aware module is removed. This performance degradation underscores the crucial role this module plays in preserving model accuracy. Its absence results in the model’s inability to effectively manage scale variations across different environments, directly affecting its robustness and its capacity to accurately estimate homography.
To evaluate the effect of fusing semantic features, we excluded the SEM and SMC modules, using only the TAHEM for training. The experimental results, also presented in the same figures, show that, Compared with our method, it is true that the methods of SMC and SEM modules do not decrease significantly in the three indicators of PSNR, SSIM and NCC, but the indicators of MACE and PME are significantly worse, and the model without these two modules has significantly worse indicators. More specifically, these results highlight that networks incorporating SEM and SMC outperform those relying solely on TAHEM, emphasizing the vital role that semantic information plays in homography estimation.

The results of the ablation test regarding the network structure are shown in Fig.~\ref{fig:tran}. It can be seen that without Hierarchical Scale-aware network, the results of the homography estimation will show a large deviation. However, the matching effect without semantic guidance module is worse than SeFNet only in the details of homography matching, which reflects that the semantic network in our model can adjust the edge part of the homography estimation results, so as to achieve better matching effect.

To verify whether semantic information is implicitly improved, thus narrowing the gap between semantic and structural information, we extract features from models with and without meta-learning strategies under four different environmental conditions, and visualize the results using heat maps. As shown in Fig.~\ref{fig:heapmap}, the visualization results of the heat map clearly demonstrate the effect of feature extraction.

With regard to the effect of meta-learning strategies on feature clarity, we observed that in a relatively simple "normal" environment, there was little difference in the extracted features between with and without meta-learning strategies. However, a closer look reveals that when meta-learning strategies are applied, the extracted features exhibit finer details. In more complex and challenging environments, meta-learning strategies enhance features more significantly. Under the harsh conditions, the features of the visualized heat map without using the meta-learning strategy are fuzzy and rough. The heat map of our model shows more obvious and detailed features. These more obvious features will help the model to understand the relationship between the images improve the accuracy of homography estimation. Thus, semantic information can better adapt to changes in different environments under the application of meta-learning strategies, which provides strong support for achieving stronger generalization ability in complex real-world applications in the future.

\section{Conclusion}

In this paper, we introduce SeFENet, a network designed to enhance homography estimation in harsh environments through the integration of semantic features. Unlike previous methods that depend solely on structural features, SeFENet uses a pretrained SEM to capture semantic features from images captured in harsh conditions. These semantic features are fused with structural features at corresponding scales within the SMC, leading to improved performance in TAHEM. Experimental results show that our method achieves robust and highly accurate homography estimation on both synthetic and real-world harsh environment datasets.

\bibliographystyle{ieeetr} 
\bibliography{main}

\begin{thebibliography}{10}

\bibitem{brown2007automatic}
M.~Brown and D.~G. Lowe, ``Automatic panoramic image stitching using invariant features,'' {\em International journal of computer vision}, vol.~74, pp.~59--73, 2007.

\bibitem{gao2011constructing}
J.~Gao, S.~J. Kim, and M.~S. Brown, ``Constructing image panoramas using dual-homography warping,'' in {\em CVPR 2011}, pp.~49--56, IEEE, 2011.

\bibitem{guo2016joint}
H.~Guo, S.~Liu, T.~He, S.~Zhu, B.~Zeng, and M.~Gabbouj, ``Joint video stitching and stabilization from moving cameras,'' {\em IEEE Transactions on Image Processing}, vol.~25, no.~11, pp.~5491--5503, 2016.

\bibitem{brady2012multiscale}
D.~J. Brady, M.~E. Gehm, R.~A. Stack, D.~L. Marks, D.~S. Kittle, D.~R. Golish, E.~Vera, and S.~D. Feller, ``Multiscale gigapixel photography,'' {\em Nature}, vol.~486, no.~7403, pp.~386--389, 2012.

\bibitem{shao2021localtrans}
R.~Shao, G.~Wu, Y.~Zhou, Y.~Fu, L.~Fang, and Y.~Liu, ``Localtrans: A multiscale local transformer network for cross-resolution homography estimation,'' in {\em Proceedings of the IEEE/CVF international conference on computer vision}, pp.~14890--14899, 2021.

\bibitem{ying2021unaligned}
J.~Ying, H.-L. Shen, and S.-Y. Cao, ``Unaligned hyperspectral image fusion via registration and interpolation modeling,'' {\em IEEE Transactions on Geoscience and Remote Sensing}, vol.~60, pp.~1--14, 2021.

\bibitem{zhou2019integrated}
Y.~Zhou, A.~Rangarajan, and P.~D. Gader, ``An integrated approach to registration and fusion of hyperspectral and multispectral images,'' {\em IEEE Transactions on Geoscience and Remote Sensing}, vol.~58, no.~5, pp.~3020--3033, 2019.

\bibitem{zhan2022homography}
X.~Zhan, Y.~Liu, J.~Zhu, and Y.~Li, ``Homography decomposition networks for planar object tracking,'' in {\em Proceedings of the AAAI Conference on Artificial Intelligence}, vol.~36, pp.~3234--3242, 2022.

\bibitem{zhang2022hvc}
H.~Zhang and Y.~Ling, ``Hvc-net: Unifying homography, visibility, and confidence learning for planar object tracking,'' in {\em European Conference on Computer Vision}, pp.~701--718, Springer, 2022.

\bibitem{mur2015orb}
R.~Mur-Artal, J.~M.~M. Montiel, and J.~D. Tardos, ``Orb-slam: a versatile and accurate monocular slam system,'' {\em IEEE transactions on robotics}, vol.~31, no.~5, pp.~1147--1163, 2015.

\bibitem{engel2014lsd}
J.~Engel, T.~Sch{\"o}ps, and D.~Cremers, ``Lsd-slam: Large-scale direct monocular slam,'' in {\em European conference on computer vision}, pp.~834--849, Springer, 2014.

\bibitem{goforth2019gps}
H.~Goforth and S.~Lucey, ``Gps-denied uav localization using pre-existing satellite imagery,'' in {\em 2019 International conference on robotics and automation (ICRA)}, pp.~2974--2980, IEEE, 2019.

\bibitem{zhao2021deep}
Y.~Zhao, X.~Huang, and Z.~Zhang, ``Deep lucas-kanade homography for multimodal image alignment,'' in {\em Proceedings of the IEEE/CVF conference on computer vision and pattern recognition}, pp.~15950--15959, 2021.

\bibitem{detone2016deep}
D.~DeTone, T.~Malisiewicz, and A.~Rabinovich, ``Deep image homography estimation,'' {\em arXiv preprint arXiv:1606.03798}, 2016.

\bibitem{erlik2017homography}
F.~Erlik~Nowruzi, R.~Laganiere, and N.~Japkowicz, ``Homography estimation from image pairs with hierarchical convolutional networks,'' in {\em Proceedings of the IEEE international conference on computer vision workshops}, pp.~913--920, 2017.

\bibitem{le2020deep}
H.~Le, F.~Liu, S.~Zhang, and A.~Agarwala, ``Deep homography estimation for dynamic scenes,'' in {\em Proceedings of the IEEE/CVF conference on computer vision and pattern recognition}, pp.~7652--7661, 2020.

\bibitem{zhang2020content}
J.~Zhang, C.~Wang, S.~Liu, L.~Jia, N.~Ye, J.~Wang, J.~Zhou, and J.~Sun, ``Content-aware unsupervised deep homography estimation,'' in {\em Computer Vision--ECCV 2020: 16th European Conference, Glasgow, UK, August 23--28, 2020, Proceedings, Part I 16}, pp.~653--669, Springer, 2020.

\bibitem{zhu2024mcnet}
H.~Zhu, S.-Y. Cao, J.~Hu, S.~Zuo, B.~Yu, J.~Ying, J.~Li, and H.-L. Shen, ``Mcnet: Rethinking the core ingredients for accurate and efficient homography estimation,'' in {\em Proceedings of the IEEE/CVF Conference on Computer Vision and Pattern Recognition}, pp.~25932--25941, 2024.

\bibitem{Wang_Liu_Zhang_Xu_Wang_2024}
Y.~Wang, H.~Liu, C.~Zhang, L.~Xu, and Q.~Wang, ``Mask-homo: Pseudo plane mask-guided unsupervised multi-homography estimation,'' {\em Proceedings of the AAAI Conference on Artificial Intelligence}, vol.~38, pp.~5678--5685, Mar. 2024.

\bibitem{jiang2023semi}
H.~Jiang, H.~Li, Y.~Lu, S.~Han, and S.~Liu, ``Semi-supervised deep large-baseline homography estimation with progressive equivalence constraint,'' in {\em Proceedings of the AAAI Conference on Artificial Intelligence}, vol.~37, pp.~1024--1032, 2023.

\bibitem{jiang2023supervised}
H.~Jiang, H.~Li, S.~Han, H.~Fan, B.~Zeng, and S.~Liu, ``Supervised homography learning with realistic dataset generation,'' in {\em Proceedings of the IEEE/CVF International Conference on Computer Vision}, pp.~9806--9815, 2023.

\bibitem{lowe2004sift}
G.~Lowe, ``Sift-the scale invariant feature transform,'' {\em Int. J}, vol.~2, no.~91-110, p.~2, 2004.

\bibitem{bay2006surf}
H.~Bay, T.~Tuytelaars, and L.~Van~Gool, ``Surf: Speeded up robust features,'' in {\em Computer Vision--ECCV 2006: 9th European Conference on Computer Vision, Graz, Austria, May 7-13, 2006. Proceedings, Part I 9}, pp.~404--417, Springer, 2006.

\bibitem{rublee2011orb}
E.~Rublee, V.~Rabaud, K.~Konolige, and G.~Bradski, ``Orb: An efficient alternative to sift or surf,'' in {\em 2011 International conference on computer vision}, pp.~2564--2571, Ieee, 2011.

\bibitem{bian2017gms}
J.~Bian, W.-Y. Lin, Y.~Matsushita, S.-K. Yeung, T.-D. Nguyen, and M.-M. Cheng, ``Gms: Grid-based motion statistics for fast, ultra-robust feature correspondence,'' in {\em Proceedings of the IEEE conference on computer vision and pattern recognition}, pp.~4181--4190, 2017.

\bibitem{suarez2020beblid}
I.~Su{\'a}rez, G.~Sfeir, J.~M. Buenaposada, and L.~Baumela, ``Beblid: Boosted efficient binary local image descriptor,'' {\em Pattern recognition letters}, vol.~133, pp.~366--372, 2020.

\bibitem{detone2018superpoint}
D.~DeTone, T.~Malisiewicz, and A.~Rabinovich, ``Superpoint: Self-supervised interest point detection and description,'' in {\em Proceedings of the IEEE conference on computer vision and pattern recognition workshops}, pp.~224--236, 2018.

\bibitem{sun2021loftr}
J.~Sun, Z.~Shen, Y.~Wang, H.~Bao, and X.~Zhou, ``Loftr: Detector-free local feature matching with transformers,'' in {\em Proceedings of the IEEE/CVF conference on computer vision and pattern recognition}, pp.~8922--8931, 2021.

\bibitem{sarlin2020superglue}
P.-E. Sarlin, D.~DeTone, T.~Malisiewicz, and A.~Rabinovich, ``Superglue: Learning feature matching with graph neural networks,'' in {\em Proceedings of the IEEE/CVF conference on computer vision and pattern recognition}, pp.~4938--4947, 2020.

\bibitem{derpanis2010overview}
K.~G. Derpanis, ``Overview of the ransac algorithm,'' {\em Image Rochester NY}, vol.~4, no.~1, pp.~2--3, 2010.

\bibitem{barath2019magsac}
D.~Barath, J.~Matas, and J.~Noskova, ``Magsac: marginalizing sample consensus,'' in {\em Proceedings of the IEEE/CVF conference on computer vision and pattern recognition}, pp.~10197--10205, 2019.

\bibitem{lu2022video}
L.~Lu, R.~Wu, H.~Lin, J.~Lu, and J.~Jia, ``Video frame interpolation with transformer,'' in {\em Proceedings of the IEEE/CVF Conference on Computer Vision and Pattern Recognition}, pp.~3532--3542, 2022.

\bibitem{ye2021motion}
N.~Ye, C.~Wang, H.~Fan, and S.~Liu, ``Motion basis learning for unsupervised deep homography estimation with subspace projection,'' in {\em Proceedings of the IEEE/CVF international conference on computer vision}, pp.~13117--13125, 2021.

\bibitem{hong2022unsupervised}
M.~Hong, Y.~Lu, N.~Ye, C.~Lin, Q.~Zhao, and S.~Liu, ``Unsupervised homography estimation with coplanarity-aware gan,'' in {\em Proceedings of the IEEE/CVF conference on computer vision and pattern recognition}, pp.~17663--17672, 2022.

\bibitem{liu2024semanticawarerepresentationlearninghomography}
Y.~Liu, Q.~Huang, S.~Hui, J.~Fu, S.~Zhou, K.~Wu, P.~Li, and J.~Wang, ``Semantic-aware representation learning for homography estimation,'' 2024.

\bibitem{Wang_2019_ICCV}
Y.-X. Wang, D.~Ramanan, and M.~Hebert, ``Meta-learning to detect rare objects,'' in {\em Proceedings of the IEEE/CVF International Conference on Computer Vision (ICCV)}, October 2019.

\bibitem{Elsken_2020_CVPR}
T.~Elsken, B.~Staffler, J.~H. Metzen, and F.~Hutter, ``Meta-learning of neural architectures for few-shot learning,'' in {\em Proceedings of the IEEE/CVF Conference on Computer Vision and Pattern Recognition (CVPR)}, June 2020.

\bibitem{Volpi_2021_CVPR}
R.~Volpi, D.~Larlus, and G.~Rogez, ``Continual adaptation of visual representations via domain randomization and meta-learning,'' in {\em Proceedings of the IEEE/CVF Conference on Computer Vision and Pattern Recognition (CVPR)}, pp.~4443--4453, June 2021.

\bibitem{Min_2023_WACV}
C.~Min, T.~Kim, and J.~Lim, ``Meta-learning for adaptation of deep optical flow networks,'' in {\em Proceedings of the IEEE/CVF Winter Conference on Applications of Computer Vision (WACV)}, pp.~2145--2154, January 2023.

\bibitem{zhang2024metadiff}
B.~Zhang, C.~Luo, D.~Yu, X.~Li, H.~Lin, Y.~Ye, and B.~Zhang, ``Metadiff: Meta-learning with conditional diffusion for few-shot learning,'' in {\em Proceedings of the AAAI Conference on Artificial Intelligence}, vol.~38, pp.~16687--16695, 2024.

\bibitem{konwer2023enhancing}
A.~Konwer, X.~Hu, J.~Bae, X.~Xu, C.~Chen, and P.~Prasanna, ``Enhancing modality-agnostic representations via meta-learning for brain tumor segmentation,'' in {\em Proceedings of the IEEE/CVF International Conference on Computer Vision}, pp.~21415--21425, 2023.

\bibitem{wang2023improving}
L.~Wang, S.~Zhou, S.~Zhang, X.~Chu, H.~Chang, and W.~Zhu, ``Improving generalization of meta-learning with inverted regularization at inner-level,'' in {\em Proceedings of the IEEE/CVF Conference on Computer Vision and Pattern Recognition}, pp.~7826--7835, 2023.

\bibitem{9189802}
T.~Xue and H.~Yu, ``Model-agnostic metalearning-based text-driven visual navigation model for unfamiliar tasks,'' {\em IEEE Access}, vol.~8, pp.~166742--166752, 2020.

\bibitem{9174763}
J.~Li, S.~Shang, and L.~Chen, ``Domain generalization for named entity boundary detection via metalearning,'' {\em IEEE Transactions on Neural Networks and Learning Systems}, vol.~32, no.~9, pp.~3819--3830, 2021.

\bibitem{PAMBALA202193}
A.~K. Pambala, T.~Dutta, and S.~Biswas, ``Sml: Semantic meta-learning for few-shot semantic segmentation,'' {\em Pattern Recognition Letters}, vol.~147, pp.~93--99, 2021.

\bibitem{gong2021cluster}
R.~Gong, Y.~Chen, D.~P. Paudel, Y.~Li, A.~Chhatkuli, W.~Li, D.~Dai, and L.~Van~Gool, ``Cluster, split, fuse, and update: Meta-learning for open compound domain adaptive semantic segmentation,'' in {\em Proceedings of the IEEE/CVF conference on computer vision and pattern recognition}, pp.~8344--8354, 2021.

\bibitem{li2021fewshotmetalearningpointcloud}
X.~Li, L.~Feng, L.~Li, and C.~Wang, ``Few-shot meta-learning on point cloud for semantic segmentation,'' 2021.

\bibitem{zhang2021meta}
G.~Zhang, Z.~Luo, K.~Cui, and S.~Lu, ``Meta-detr: Few-shot object detection via unified image-level meta-learning,'' {\em arXiv preprint arXiv:2103.11731}, vol.~2, no.~6, p.~2, 2021.

\bibitem{redmon2016you}
J.~Redmon, S.~Divvala, R.~Girshick, and A.~Farhadi, ``You only look once: Unified, real-time object detection,'' in {\em Proceedings of the IEEE conference on computer vision and pattern recognition}, pp.~779--788, 2016.

\bibitem{liu2016ssd}
W.~Liu, D.~Anguelov, D.~Erhan, C.~Szegedy, S.~Reed, C.-Y. Fu, and A.~C. Berg, ``Ssd: Single shot multibox detector,'' in {\em Computer Vision--ECCV 2016: 14th European Conference, Amsterdam, The Netherlands, October 11--14, 2016, Proceedings, Part I 14}, pp.~21--37, Springer, 2016.

\bibitem{wu2022edge}
Y.~Wu, H.~Guo, C.~Chakraborty, M.~R. Khosravi, S.~Berretti, and S.~Wan, ``Edge computing driven low-light image dynamic enhancement for object detection,'' {\em IEEE Transactions on Network Science and Engineering}, vol.~10, no.~5, pp.~3086--3098, 2022.

\bibitem{yuan2024plug}
J.~Yuan, Y.~Hu, Y.~Sun, B.~Wang, and B.~Yin, ``A plug-and-play image enhancement model for end-to-end object detection in low-light condition,'' {\em Multimedia Systems}, vol.~30, no.~1, p.~27, 2024.

\bibitem{zheng2022video}
Y.~Zheng and H.~Zhang, ``Video analysis in sports by lightweight object detection network under the background of sports industry development,'' {\em Computational Intelligence and Neuroscience}, vol.~2022, no.~1, p.~3844770, 2022.

\bibitem{wang2025omnitracker}
J.~Wang, Z.~Wu, D.~Chen, C.~Luo, X.~Dai, L.~Yuan, and Y.-G. Jiang, ``Omnitracker: Unifying visual object tracking by tracking-with-detection,'' {\em IEEE Transactions on Pattern Analysis and Machine Intelligence}, 2025.

\bibitem{zhang2022bytetrack}
Y.~Zhang, P.~Sun, Y.~Jiang, D.~Yu, F.~Weng, Z.~Yuan, P.~Luo, W.~Liu, and X.~Wang, ``Bytetrack: Multi-object tracking by associating every detection box,'' in {\em European conference on computer vision}, pp.~1--21, Springer, 2022.

\bibitem{alotaibi2022computational}
M.~F. Alotaibi, M.~Omri, S.~Abdel-Khalek, E.~Khalil, and R.~F. Mansour, ``Computational intelligence-based harmony search algorithm for real-time object detection and tracking in video surveillance systems,'' {\em Mathematics}, vol.~10, no.~5, p.~733, 2022.

\bibitem{cai2021yolov4}
Y.~Cai, T.~Luan, H.~Gao, H.~Wang, L.~Chen, Y.~Li, M.~A. Sotelo, and Z.~Li, ``Yolov4-5d: An effective and efficient object detector for autonomous driving,'' {\em IEEE Transactions on Instrumentation and Measurement}, vol.~70, pp.~1--13, 2021.

\bibitem{gaikwad2024identification}
D.~Gaikwad, A.~Sejal, S.~Bagade, N.~Ghodekar, and S.~Labade, ``Identification of cervical spine fracture using deep learning,'' {\em Australian Journal of Multi-Disciplinary Engineering}, vol.~20, no.~1, pp.~48--56, 2024.

\bibitem{salman2022automated}
M.~E. Salman, G.~{\c{C}}. {\c{C}}akar, J.~Azimjonov, M.~K{\"o}sem, and {\.I}.~H. Cedio{\u{g}}lu, ``Automated prostate cancer grading and diagnosis system using deep learning-based yolo object detection algorithm,'' {\em Expert Systems with Applications}, vol.~201, p.~117148, 2022.

\bibitem{cao2023recurrent}
S.-Y. Cao, R.~Zhang, L.~Luo, B.~Yu, Z.~Sheng, J.~Li, and H.-L. Shen, ``Recurrent homography estimation using homography-guided image warping and focus transformer,'' in {\em Proceedings of the IEEE/CVF Conference on Computer Vision and Pattern Recognition}, pp.~9833--9842, 2023.

\bibitem{everingham2010pascal}
M.~Everingham, L.~Van~Gool, C.~K. Williams, J.~Winn, and A.~Zisserman, ``The pascal visual object classes (voc) challenge,'' {\em International journal of computer vision}, vol.~88, pp.~303--338, 2010.

\bibitem{wang2004image}
Z.~Wang, A.~C. Bovik, H.~R. Sheikh, and E.~P. Simoncelli, ``Image quality assessment: from error visibility to structural similarity,'' {\em IEEE transactions on image processing}, vol.~13, no.~4, pp.~600--612, 2004.

\bibitem{zitova2003image}
B.~Zitova and J.~Flusser, ``Image registration methods: a survey,'' {\em Image and vision computing}, vol.~21, no.~11, pp.~977--1000, 2003.

\bibitem{lowe2004distinctive}
D.~G. Lowe, ``Distinctive image features from scale-invariant keypoints,'' {\em International journal of computer vision}, vol.~60, pp.~91--110, 2004.

\bibitem{fischler1981random}
M.~A. Fischler and R.~C. Bolles, ``Random sample consensus: a paradigm for model fitting with applications to image analysis and automated cartography,'' {\em Communications of the ACM}, vol.~24, no.~6, pp.~381--395, 1981.

\end{thebibliography}


\vfill
\end{document}